\pdfoutput=1
\documentclass[11pt]{article}

\usepackage[preprint]{acl}

\usepackage{times}
\usepackage{latexsym}

\usepackage[T1]{fontenc}
\usepackage[utf8]{inputenc}

\usepackage{microtype}

\usepackage{inconsolata}

\usepackage{graphicx}

\usepackage{url}
\usepackage{xurl}
\usepackage[T1]{fontenc}
\usepackage{amsmath}
\usepackage{graphicx}
\usepackage{booktabs}
\usepackage{multirow}
\usepackage{pifont}
\usepackage{pgfplots}
\usepackage{tabularx}
\usepackage{subcaption}
\usepackage[edges]{forest}
\usepackage[acronym,nonumberlist,nopostdot]{glossaries}
\usepackage{enumitem}
\setlist{leftmargin=5.5mm}
\usepackage[skins,breakable]{tcolorbox}
\usepackage{cleveref}

\usepackage{tikz}
\usetikzlibrary{arrows.meta}  % optional, for arrow customization

\glsdisablehyper  % Disable links from glossary entries
\makeglossaries

\newacronym{uplead}{\texttt{U-PLEAD}}{\texttt{Unseen-PLEAD}}
\newacronym{genbench}{\texttt{TARGET}}{Testing Atomic Reasoning in Generalisation of Expressions and Targets}

\usetikzlibrary{arrows.meta, positioning}
\pgfplotsset{compat=1.18}

\definecolor{slot1color}{RGB}{70, 130, 180}   % Steel Blue
\definecolor{slot2color}{RGB}{255, 140, 0}    % Dark Orange
\definecolor{slot3color}{RGB}{112, 128, 144}  % Slate Gray

\title{Compositional Generalisation for Explainable Hate Speech Detection}

\author{
 \textbf{Agostina Calabrese\textsuperscript{1}},
 \textbf{Tom Sherborne\textsuperscript{2}},
 \textbf{Bj{\"o}rn Ross\textsuperscript{1}},
 \textbf{Mirella Lapata\textsuperscript{1}}
\\
 \textsuperscript{1}School of Informatics, University of Edinburgh,
 \textsuperscript{2}Cohere,
\\
 \small{
   a.calabrese@ed.ac.uk
 }
}

\begin{document}
\maketitle
\begin{abstract}
Hate speech detection is key to online content moderation, but current models struggle to generalise beyond their training data.
This has been linked to dataset biases and the use of sentence-level labels, which fail to teach models the  underlying structure of hate speech.
In this work, we show that even when models are trained with more fine-grained, span-level annotations (e.g., ``artists'' is labeled as  \texttt{target} and ``are parasites'' as \texttt{dehumanising comparison}), they struggle to disentangle the meaning of these labels from the surrounding context. As a result, combinations of expressions that deviate from those seen during training remain particularly difficult for models to detect.
We investigate whether training on a dataset where expressions occur with equal frequency across all contexts can improve generalisation.
To this end, we create  \gls{uplead}, a dataset of ${\sim}364{,}000$ synthetic posts, along with a novel compositional generalisation benchmark of ${\sim}8{,}000$ manually validated posts. Training on a combination of \texttt{U-PLEAD} and real data improves compositional generalisation while achieving state-of-the-art performance on the human-sourced PLEAD.
\end{abstract}

\section{Introduction}
%Recent developments in the 2024 US election, coupled with Elon Musk\footnote{\url{https://www.politico.eu/article/elon-musks-x-eu-safe-space-free-speech-digital-services-act/}} and Mark Zuckerberg's\footnote{\url{https://www.nbcnews.com/tech/social-media/meta-new-hate-speech-rules-allow-users-call-lgbtq-people-mentally-ill-rcna186700}} controversial views on content moderation, have made research on hate speech detection feel both outdated and more urgent than ever. 
A large body of research has focused on developing models for the automatic detection of online hate speech (e.g., \citealt{2012-warner-detecting,2019-mozafari-bert,2021-saeidi-trees}), but their effectiveness has been largely overestimated due to evaluations conducted on academic datasets. \citet{2024-tonneau-hateday}  show that models fail to generalise to datasets with different target distributions, like the one found in a 24-hour period Twitter stream. \citet{2022-calabrese-plead} argue that the problem stems from the fact the task is typically framed as a binary sequence classification problem: datasets where posts are assigned a \emph{single} sentence-level label are inadequate for learning the concept of {hate speech}. They demonstrate that the same annotations can correspond to multiple underlying phenomena, and that even small changes such as different random initializations can cause the same model to learn entirely different patterns. This motivates their use of span-level annotations, where different \textit{slot} labels are used to indicate a \texttt{target}, the mention of a \texttt{protected characteristic}, or a \texttt{threat} in a post. Slot labels are supposed to guide models toward learning policy-relevant phenomena, assuming that  these labels are interpreted as atomic properties irrespective of their surrounding context. %But is this really the case? In this work we show that models trained for policy-aware abuse detection 

In this paper,  we show that models trained with slot annotations do not always grasp their atomicity and remain vulnerable to unintended correlations in the training dataset.
%Where do these undesired correlations come from? A documented source is selection bias \cite{2019-wiegand-selectionbias}, as the heuristics used to gather hateful posts introduce topic bias across classes. A second source are stereotypes, power dynamics and historical events.
For instance, although comparisons with ``terrorists'' are often targeted against the Muslim community %, while references to the Black Lives Matter movement are associated to anti-Black hate speech
\cite{2022-yoder-stereotypes}, a detection model should recognise that equating any group with terrorists is inherently derogatory. However, if such expressions are never observed with different targets, slot labels alone may not be sufficient for the model to generalise this understanding. In other words, we identify that hate speech detection faces the same \textit{compositional generalisation} challenges (i.e.,~generalisation to unseen combinations of known phrases) observed in other span-based NLU tasks such as semantic parsing \citep{zheng-lapata-2021-compositional-generalization,2023-hupkes-generalisation}.

While these correlations are unavoidable in naturally occurring data due to the existence of stereotypes and power dynamics, we generate a collection of synthetic posts designed to be free from this issue by balancing the frequency of label combinations. Our hypothesis is:~this synthetic data would help models disentangle the meaning of slot labels from their surrounding context, and therefore generalise better to unseen distributions.
% We hypothesise that training a model on a resource where the behaviour of each expression is fully represented---at the cost of limited linguistic diversity---will improve its generalisation capabilities. 
We start from the structured hate speech definition and annotations provided in {PLEAD} \cite{2022-calabrese-plead} and %define a grammar $G$. We use $G$ and 
use Large Language Models (LLMs) to generate \gls{uplead}, a dataset of ${\sim}364{,}000$ synthetic posts with no correlations between spans and classes, or targets and expressions. For instance, the derogatory expression \textit{``are terrorists''} and its equivalents appear in our dataset with the same frequency for all protected \emph{and} non-protected targets. Likewise, these expressions occur equally often in posts labelled as ``derogatory'' to those assigned to any other class. %Additionally, we generate 8 generalisation tests designed to contain unseen expressions or expression combinations. Each test counts over $1,000$ posts, and the quality is ensured through manual human validation. 

The generalisation capabilities of hate speech models are most commonly tested in an unstructured manner:~evaluating on datasets different from the training corpus or harder splits of the same dataset without a clear measure of the distribution shift between the two \cite{2021-yin-generalisation,2023-zufle-latent}. In this work, we introduce~\texttt{TARGET} (Testing Atomic Reasoning in Generalisation of Expressions and Targets). \texttt{TARGET} contains only expressions, or expression combinations, unseen in \gls{uplead}. We design 8 generalisation tests, aiming to detect cases like previously unseen hate speech targets, each of which is represented by ${\sim}1{,}000$ manually validated posts. 
 Our experiments show that partially substituting \texttt{PLEAD}'s training set with \gls{uplead} enhances the generalisation capabilities of classification \emph{and} slot-filling models without any loss in performance on the \texttt{PLEAD} test set.
%
%design \gls{genbench}, a collection of 8 generalisation tests that contain only expressions or expression combinations unseen in \gls{uplead}. Each test counts over $1{,}000$ manually validated posts and is designed to assess a specific generalisation skill, such as the ability to detect hate speech directed at previously unseen targets. We observe that augmenting \texttt{PLEAD}'s training set with \gls{uplead} enhances the generalisation capabilities of both classification and slot-filling models without any loss in performance on the \texttt{PLEAD} test set.
%
We summarise our contributions as follows:
\begin{itemize}[noitemsep]
    \item  We study compositional generalisation in the context of hate speech and develop a procedure of generating balanced synthetic posts, which we show enhance model generalisation to unseen expressions. 
    \item We create \texttt{TARGET}, the first benchmark for assessing the generalisation capabilities of hate speech models. 
    \item Through extensive experiments, we demonstrate that data augmentation, albeit with synthetic examples, improves generalisation without compromising in-domain performance. 
\end{itemize}

\section{Related Work}

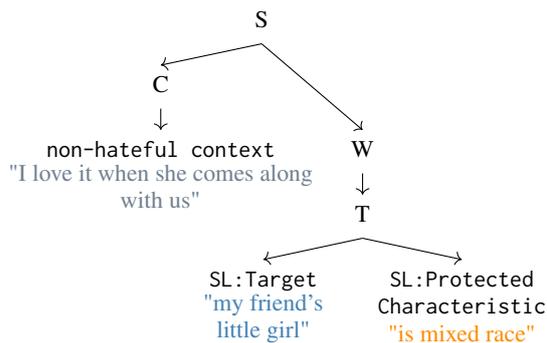
\begin{figure}
    \centering
    \noindent
    \small{Original post:\textit{ ``\textcolor{slot1color}{My friends little girl} \textcolor{slot2color}{is mixed race}, \textcolor{slot3color}{I love it when she comes along with us}.''}}\\
    \vspace{0.8em}
   \begin{forest}
      for tree={
        edge={->},
        parent anchor=south,
        child anchor=north,
        align=center,
        l sep=2pt,
        s sep=8pt,
        font=\small,
      }
      [S, name=Snode
        [C
          [\texttt{non-hateful context}\\
           \parbox{4cm}{\centering\textcolor{slot3color}{"I love it when she comes along with us"}}]
        ]
        [\phantom{}, no edge
            [W, name=Wnode, no edge
              [T
                [\texttt{SL:Target}\\
                 \parbox{2cm}{\centering\textcolor{slot1color}{"my friend's little girl"}}]
                [\texttt{SL:Protected}\\\texttt{Characteristic}\\
                 \parbox{2cm}{\centering\textcolor{slot2color}{"is mixed race"}}]
              ]
            ]
        ]
        edge path={
        \draw[\noexpand->] (Snode.south) to (Wnode.north);
        }
      ]
    \end{forest}
    % }
    \vspace{-.4cm}
\caption{Parse tree generation for a PLEAD post  using our Hate Speech Grammar.}
\label{fig:grammar}
\end{figure}
Hate speech is content that targets individuals or groups on the basis of their protected characteristics (e.g., gender) with derogatory language (explicit or implicit), dehumanising comparisons, and threatening language. Content that explicitly glorifies or supports hateful events or organizations is also considered hate speech \cite{2021-vidgen-lftw}. 

Generalisation and robustness have been persistent weaknesses throughout the history of hate speech research \cite{2019-wiegand-selectionbias,2020-kennedy-debias,2023-reyero-multidisciplinary}. While the advent  of larger models and  promising results from training on dynamically generated adversarial data \cite{2021-vidgen-lftw} suggested  the problem might be solved, at least in English, \citet{2024-tonneau-hateday} showed that these models still struggle to generalise to unseen distributions. %This cycle of endlessly collecting more data to fix shortcomings, only to later expose new ones, is not a sustainable solution.
\citet{2022-calabrese-plead} argued that 
%for models to stop blindly searching for patterns, that often turn out to be incorrect, 
models require more information about specific hate speech phenomena they are meant to detect. To achieve this, they distill a hate speech policy into atomic properties (i.e.,~slots), and conceptualise  the task as an instance of intent classification and slot filling. In this setting, models are not just shown that \textit{``Artists are parasites''} is not hateful, but also receive additional information: \textit{``Artists''} is the \texttt{target}, \textit{``are parasites''} is a \texttt{dehumanising comparison}, and 
the post cannot be considered hateful based on these two slots, as it lacks any mention of protected characteristics. 

The decomposition into slots and intents assumes models will grasp the atomicity of the slots and  disentangle their meaning from the surrounding context. However, protected groups are targeted in distinct, specific ways, and comparisons to parasites are more commonly associated with groups like immigrants than, e.g.,~women \cite{2012-haas-stereotype}. As a result, models may learn that \textit{``are parasites''} functions as a dehumanising comparison only when paired with terms like \textit{``Artists''} or \textit{``Immigrants''}, but not as a generalisable pattern. This tendency is further amplified by neural network models, which often default to relying on extra, unnecessary features instead of learning the minimal features needed to define a category boundary.
\cite{2022-dasgupta-exemplar}.

In this work we focus on finding a more robust training approach that enables a model to generalise to any unseen distribution, rather than designing a technique for adapting a model to a specific new distribution \cite{2022-sarwar-augmentation}. We use data augmentation to improve generalisation, however, unlike \citet{2021-mostafazadeh-counterfactual}, we intentionally include implausible posts in our synthetic data, as this is crucial for eliminating undesired correlations.
We propose several tests for compositional generalisation in the hate speech domain. 
We relax the classical definition of compositional generalisation which assumes  that all expressions appearing in the test set are also seen during training, with only their combinations being novel (e.g., \citealt{2023-lindemann-compgen}). By allowing expressions to appear only at test time, we can evaluate more domain-relevant generalisation scenarios, such as those introduced by policy changes (e.g.,~recognising pregnancy as a protected characteristic) or new social events (e.g.,~COVID-19-related hate targeting Asians), and avoid contamination. Recent studies \cite{2022-kim-lexical} have shown that assuming expressions are ``unknown'' solely because they are absent from the training set overstimates the generalisation capabilities of pre-trained language models.

\section{Compositional Generalisation via Data Augmentation}

We aim to create a dataset consisting of a $\langle$training, test$\rangle$ set pair such that it allows us to test the hypothesis that training a model on a resource with \emph{exhaustive} coverage of an expression’s behaviour will improve generalisation.
Traditional compositional generalisation settings require the test set to contain \emph{unseen} combinations of \emph{seen} expressions. In the context of hate speech, this implies that the behaviour of these expressions is not fully represented in the training data, and  that we are testing combinations of targets and expressions that may not naturally co-occur in real-world hate speech, potentially focusing on unrealistic examples. Instead, we leverage the broad pre-training of modern language models to relax the constraint that \emph{all} spans must appear in the training data. We generate a training set that provides  exhaustive coverage for each expression and design a set of test cases that challenge models to recognise known expressions or targets in novel contexts. 
We next elaborate on how this data is generated.

\subsection{The Hate Speech Grammar}

To study compositional generalisation for hate speech, we need to define the possible targets, as well as hateful and non-hateful expressions, and determine how they can be combined. We take advantage of the annotations in the PLEAD dataset (Appendix~\ref{sec:plead}) which were designed to generate explanations, associating each post with a tree-like structure where  leaves represent a span of tokens in a post,   internal nodes correspond to slot labels, and the root to intent labels (see Figure~\ref{fig:grammar}). Their ontology includes the following slot labels: \texttt{target} and \texttt{protected characteristic} ($T$), \texttt{dehumanising comparison} ($D$), \texttt{threatening speech} ($T_h$), \texttt{negative} and \texttt{derogatory opinion} ($N$), \texttt{hate entity} ($E$), \texttt{support of hate crimes} ($S$), and \texttt{negative stance} ($N_s$). 

Based on their ontology, we  define a formal grammar~$G$ that can generate trees associated with hateful and non-hateful posts. Any span of text with a slot label in PLEAD (e.g., the target \textit{``Those women''} or the negative stance \textit{``these claims are not true''}) is a terminal symbol in~$G$.  Additional terminal symbols are created by removing the target and any possible protected characteristic from non-hateful posts with no other associated slot labels. We refer to these as ``non-hateful context''~($C$). For example, we extract the non-hateful context \textit{``I love it when she comes along with us''} from the post \textit{``my friend's little girl is mixed race, I love it when she comes along with us''} where \textit{``my friend's little girl''} is tagged as target, and \textit{``is mixed race''} as protected characteristic. 

We define a non-terminal symbol for each slot in the ontology, with a few exceptions. The target and protected characteristic symbols are merged into $T$ as they are not independent, and derogatory or negative opinions are merged into $N$ due to their overlap.\footnote{The main distinction between these two slots is the degree of explicitness in expressing a derogatory opinion.} For each non-terminal symbol $N_t$ and for each terminal symbol $t$ associated to the corresponding slot, we define a production rule: $N_t \rightarrow t$. Additionally, we define the following production rules:
\begin{equation*}
  S \rightarrow C P \,|\,C W\,|\,C W P  
\end{equation*}
\begin{equation*}
  W \rightarrow \epsilon\,|\,T W\,|\,E W
\end{equation*}
\begin{equation*}
  P \rightarrow \epsilon\,|\,D P\,|\,T_h P\,|\,N P\,|\,S P\,|\,N_s P  
\end{equation*}
where $S$ is the start symbol and $\epsilon$ is the empty string. Figure~\ref{fig:grammar} illustrates how $G$ can generate the tree associated with a non-hateful post.
\subsection{The \gls{uplead} Dataset}
\label{sec:bias-free}
% To create a dataset free from stereotypes we first generate a balanced
To create the training set, we generate a collection of trees from~$G$, and ``translate'' them into posts. We generate trees based on the following criteria:
\begin{tcolorbox}
\small 
\begin{description}[noitemsep]
    \item[\hypertarget{desc:c1}{C1}] Each protected and non-protected target appears with same frequency across all classes. %classification classes.
    \item[\hypertarget{desc:c2}{C2}] Each dehumanising comparison, threat, negative opinion, and expression of support for hate crimes occurs with same relative frequency across all classes. %classification classes.
    \item[\hypertarget{desc:c3}{C3}] Each hate entity appears with same frequency across all classes. %classification classes.
    \item[\hypertarget{desc:c4}{C4}] Each negative stance expression appears with same frequency across all classes. %classification classes.
   \hspace{-1cm} \item[\hypertarget{desc:c5}{C5}] Each protected and non-protected target and each hate entity occur with each dehumanising comparison, threat, negative opinion, and expression of support for hate crimes with same frequency.
\end{description}
\end{tcolorbox}

Allowing many targets and expressions in the dataset would lead to a combinatorial explosion of the number of instances. Therefore, we limit the number of terminal symbols in $G$ used for generation. 
To maximise linguistic diversity, 
we cluster all non-protected targets, hate entities, dehumanising comparisons, threats, negative opinions, expressions of support for hate crimes, and negative stances in PLEAD based on semantic similarity. We compute a vector representation for each expression using Sentence-BERT \cite{2019-reimers-sentencebert} and then perform hierarchical clustering. %\footnote{The threshold value ($0.5$) was determined empirically by evaluating the clustering quality across different values.} 
Expressions that are assigned to the same cluster are considered equivalent (e.g., the threatening expressions \textit{``i want to burn''}, \textit{``burn to the ground''}, and \textit{``burned''}). For protected targets, we take a different approach and treat spans as equivalent if they are tagged with the same target group in the original dataset (e.g., \textit{``woman''}, \textit{``she''} and \textit{``her''} for the group ``women''). We select $40$ (clusters of) protected targets and hate entities, and $20$ clusters from other slots, to generate a balanced collection of $384{,}800$ trees. We provide more details in Appendix~\ref{sec:tree_generation}.

We take advantage of the linguistic abilities of LLMs to convert the trees generated by our grammar into posts. 
We generate a first draft  using Vicuna-30B-Uncensored, a model trained without responses containing alignment or moralising content in its pre-training corpus. %\footnote{\url{https://huggingface.co/cognitivecomputations/Wizard-Vicuna-30B-Uncensored}}
We prompt the model in a few-shot setting, and adapt the in-context examples to align with the structure of the tree for which we are generating a post. The instruction specifies which spans must occur in the post (possibly verbatim), and the role each span must play (see Appendix~\ref{sec:post-generation}). 
To improve fluency and ensure the spans are included in the posts, we refine them using GPT-3.5-Turbo. 
We discard trees for which GPT fails to generate all spans even after multiple rounds ($5.34\%$), returning the final collection of $364,261$ \texttt{<post,\,tree>} pairs. Some posts may be noisy, due to the large size of the tree or the challenge of connecting highly unrelated spans, however, the large size of the dataset compensates for this noise \cite{2022-sarwar-augmentation}. 
We will make \gls{uplead} available to the  research community upon request.%, requiring users to submit a declaration of intent.%outlining their intended use.
\subsection{The \texttt{TARGET} Benchmark}
\begin{table*}[t]
\setlength\extrarowheight{-2pt}
    \centering
    \begin{tabular}{ll|ccc|cccccl}
        \toprule
        \multirow{2}{*}{Test} & \multirow{2}{*}{Tree} & \multicolumn{3}{c|}{Unseen Span} & \multicolumn{5}{c}{Classes} & \multicolumn{1}{c}{\multirow{2}{*}{Example}} \\
         &  & \small $T|E$ & \small $X|S$ & \small $N_s$ & \small C & \small T & \small D & \small P & \small N &  \\
        \midrule
        \multirow{3}{*}{1} & \multirow{3}{*}{$T X$} & \multirow{3}{*}{\ding{51}} & \multirow{3}{*}{\ding{55}} & \multirow{3}{*}{--} & \multirow{3}{*}{\ding{51}} & \multirow{3}{*}{\ding{51}} & \multirow{3}{*}{\ding{51}} & \multirow{3}{*}{\ding{55}} & \multirow{3}{*}{\ding{51}}  & \footnotesize{\multirow{3}{*}{\shortstack[l]{I can't believe the state of this place, \emph{a trashy}\\\emph{grotesque looking house}. \emph{Why is my city is filled}\\\emph{with all kinds of plagues: rats, cockroaches, spiders}}}} \\
        &&&&&&&&&\\
        &&&&&&&&&\\
        \multirow{3}{*}{1b} & \multirow{3}{*}{$E S$} & \multirow{3}{*}{\ding{51}} & \multirow{3}{*}{\ding{55}} & \multirow{3}{*}{--} & \multirow{3}{*}{\ding{55}} & \multirow{3}{*}{\ding{55}} & \multirow{3}{*}{\ding{55}} & \multirow{3}{*}{\ding{51}} & \multirow{3}{*}{\ding{55}}  & \footnotesize{\multirow{3}{*}{\shortstack[l]{The philosopher, Xenophanes, \emph{was a renowned}\\\emph{thinker} who once said that \"if God had not intended\\for us to \emph{rape}, he would not have given us rapists.}}} \\
        &&&&&&&&&\\
        &&&&&&&&&\\
        \multirow{2}{*}{2} & \multirow{2}{*}{$T X$} & \multirow{2}{*}{\ding{55}} & \multirow{2}{*}{\ding{51}} & \multirow{2}{*}{--} & \multirow{2}{*}{\ding{51}} & \multirow{2}{*}{\ding{51}} & \multirow{2}{*}{\ding{51}} & \multirow{2}{*}{\ding{55}} & \multirow{2}{*}{\ding{51}}  & \footnotesize{\multirow{2}{*}{\shortstack[l]{\emph{Those imams} are all just trouble makers,\\\emph{all of them should just be put down}}}} \\
        &&&&&&&&&\\
        \multirow{2}{*}{2b} & \multirow{2}{*}{$E S$} & \multirow{2}{*}{\ding{55}} & \multirow{2}{*}{\ding{51}} & \multirow{2}{*}{--} & \multirow{2}{*}{\ding{55}} & \multirow{2}{*}{\ding{55}} & \multirow{2}{*}{\ding{55}} & \multirow{2}{*}{\ding{51}} & \multirow{2}{*}{\ding{55}}  & \footnotesize{\multirow{2}{*}{\shortstack[l]{I support \emph{fascist/white supremacist} because \emph{their}\\\emph{love for} their country is admirable.}}} \\
        &&&&&&&&&\\
        \multirow{2}{*}{3} & \multirow{2}{*}{$T X N_s$} & \multirow{2}{*}{\ding{55}} & \multirow{2}{*}{\ding{55}} & \multirow{2}{*}{\ding{51}} & \multirow{2}{*}{\ding{55}} & \multirow{2}{*}{\ding{55}} & \multirow{2}{*}{\ding{55}} & \multirow{2}{*}{\ding{55}} & \multirow{2}{*}{\ding{51}}  & \footnotesize{\multirow{2}{*}{\shortstack[l]{I can't believe \emph{he yelled about} \emph{f4g\$} \emph{have no soul}.\\I'm so disappointed in him.}}} \\
        &&&&&&&&&\\
        \multirow{3}{*}{3b} & \multirow{3}{*}{$E S N_s$} & \multirow{3}{*}{\ding{55}} & \multirow{3}{*}{\ding{55}} & \multirow{3}{*}{\ding{51}} & \multirow{3}{*}{\ding{55}} & \multirow{3}{*}{\ding{55}} & \multirow{3}{*}{\ding{55}} & \multirow{3}{*}{\ding{55}} & \multirow{3}{*}{\ding{51}}  & \footnotesize{\multirow{3}{*}{\shortstack[l]{I'm disappointed that you're supporting a \emph{nationalist}\\\emph{organization}. I mean, \emph{I love} my country too, but \emph{it}\\\emph{doesn't mean I'm trash talking} other nations.}}} \\
        &&&&&&&&&\\
        &&&&&&&&&\\
        \multirow{3}{*}{4} & \multirow{3}{*}{$T X$} & \multirow{3}{*}{\ding{51}} & \multirow{3}{*}{\ding{51}} & \multirow{3}{*}{--} & \multirow{3}{*}{\ding{51}} & \multirow{3}{*}{\ding{51}} & \multirow{3}{*}{\ding{51}} & \multirow{3}{*}{\ding{55}} & \multirow{3}{*}{\ding{51}}  & \footnotesize{\multirow{3}{*}{\shortstack[l]{\emph{The scout} went on a \emph{fucking raft} illegally? That's\\so unlike them! \emph{Their Mongolian ancestor} must\\be turning in their grave.}}} \\
        &&&&&&&&&\\
        &&&&&&&&&\\
        \multirow{2}{*}{4b} & \multirow{2}{*}{$E S$} & \multirow{2}{*}{\ding{51}} & \multirow{2}{*}{\ding{51}} & \multirow{2}{*}{--} & \multirow{2}{*}{\ding{55}} & \multirow{2}{*}{\ding{55}} & \multirow{2}{*}{\ding{55}} & \multirow{2}{*}{\ding{51}} & \multirow{2}{*}{\ding{55}}  & \footnotesize{\multirow{2}{*}{\shortstack[l]{Of course it's okay! \emph{greeting-master} is \emph{it's definitely}\\\emph{ok}, in fact, it's great!}}} \\
        &&&&&&&&&\\
        \bottomrule
    \end{tabular}
    \caption{Example posts and their tree structure for each generalisation test  ($X \in \{D,T_h,N\}$); we also show which slots are filled with Unseen Spans (i.e.,~spans which are not sourced from \gls{uplead}) and which Classes or intents are present in each test (Dehumanising \underline{C}omparison, \underline{T}hreatening Speech, \underline{D}erogation, \underline{P}ro-Hate Crimes and \underline{N}ot Hateful). Spans used to generate the post are shown in \emph{italics}.}
    \label{tab:gen-tests}
\end{table*}

We design eight test cases to evaluate the compositional generalisation capabilities of hate speech detection models.
While these tests serve as a true generalisation challenge for models trained on \gls{uplead}  --- where we can control which expressions appear --- they also present a difficult benchmark for models trained on other resources.
This is because our tests are not limited to most common  protected groups or stereotypical associations, but instead reflect a broader and more balanced coverage of possible targets and linguistic patterns.
We define four tests as follows:
\begin{tcolorbox}
\small
\begin{description}[noitemsep]
    \item[T1] Instances containing unseen targets, the corresponding protected characteristic (if any), and dehumanising comparisons, threats, or negative opinions sourced from \gls{uplead}. %Each instance can be generated using the production rules of $G$ from $T\,D$, $T\,T_h$ or $T\,N$.
    \item[T2] Instances containing targets sourced from \gls{uplead}, and unseen dehumanising comparisons, threats, or negative opinions. %Each instance can be generated using the production rules of $G$ from $T\,D$, $T\,T_h$ or $T\,N$.
    \item[\hypertarget{desc:t3}{T3}] Instances containing targets, dehumanising comparisons, threats, and negative opinions sourced from \gls{uplead}, and unseen negative stance expressions. %Each instance can be generated using the production rules of $G$ from $T\,D\,N_s$, $T\,T_h\,N_s$ or $T\,N\,N_s$.
    \item[T4] Instances containing unseen targets, dehumanising comparisons, threats, and negative opinions. %Each instance can be generated using the production rules of $G$ from $T\,D$, $T\,T_h$ or $T\,N$.
\end{description}
\end{tcolorbox}
We analogously define 4 additional tests \hypertarget{desc:t3b}{$Ti_b$} ($1 \leq i \leq 4$) focusing on hate entities (instead of targets) and expressions of support for hate crimes (instead of dehumanising comparisons, threats, or negative opinions). Table~\ref{tab:gen-tests} gives an overview of the eight generalisation tests.

When selecting spans that do not occur in \gls{uplead}, we sample expressions from unused clusters (Section~\ref{sec:bias-free}). In the case of protected targets, since all clusters from PLEAD have been used to generate \gls{uplead}, we select common protected characteristics absent from PLEAD (i.e.,~pregnancy, serious diseases, and veteran status) and introduce new groups for those that are covered (i.e.,~Pacific Islanders, Arab Americans, Taiwanese, Mongolians, Nepalese, Sri Lankans, Ukrainians, Hungarians, Czechs, Colombians, and Puerto Ricans). We then instruct GPT-4o\footnote{We use GPT-3.5-Turbo for large-scale generation due to its speed and lower cost, and the latest model for smaller tasks.} to generate \texttt{<target, protected characteristic>} span pairs (e.g.,~<\textit{``the couple''}, \textit{``parenting classes''}>). We manually review these  pairs  and discard any unsuitable examples. We acknowledge the reflection of cultural biases in some pairs, notably in the names associated with specific ethnicities. However, since these instances are only intended for testing generalisation, we do not consider this a major issue.

For each test, we generate~$2{,}000$~trees matching the required structure and  use Vicuna-30B-Uncensored and GPT-3.5-Turbo to generate posts following the same approach as with \gls{uplead}. We perform only one generation round with GPT-3.5-Turbo and discard any posts that do not contain the correct spans. This process results in over $1{,}000$ instances per test, a total of $10{,593}$ \texttt{<post,\,tree>} pairs. As with \gls{uplead}, we expect these posts to be somewhat noisy and not entirely fluent. To validate the use of this data for evaluation purposes,  we recruit a domain expert to assess whether: (1) a post is fluent,  (2) the assigned classification label is correct, (3) the associated tree is accurate, and (4) the slot labels are correct.
%, even if they are not properly linked within the tree. 

We inspect $100$~instances per generalisation test, and find  that $93.5\%$~of the posts are fluent and $76.25\%$~bear the correct classification label. Posts correctly reflect the corresponding tree $69.5\%$~of the time and $92.25\%$ of the posts exhibit entirely correct slot labels. GPT-3.5-Turbo attempts to convert hateful posts into non-hateful ones only in $29.30\%$ of the cases. We will also make the \texttt{TARGET} benchmark  publicly available upon request. 
%make the test sets available to the research community upon request, requiring users to submit a declaration of intent outlining their intended use.

\label{sec:seen}
\section{Experimental Evaluation}
Our experiments were designed to assess whether models struggle with compositional generalisation in the domain of hate speech and whether \gls{uplead} can help mitigate existing shortcomings.  To ensure fairness and avoid contamination, we exclude any models  involved in generating \gls{uplead} or \texttt{TARGET} from our experiments. We report results averaged over three runs with different random seeds; for parameters and evaluation metrics, see Appendix~\ref{sec:experiments-details}.

\subsection{Models Don't Treat Slots as Atomic Concepts}
\label{sec:exp-slot}
%

%
%In our first experiment, we aim to provide empirical evidence that (1) models struggle with compositional generalisation within the domain of hate speech, and (2) the \gls{uplead} dataset can help mitigate this limitation. 
In this experiment we simplify the span labelling task by focusing solely on the threatening speech slot since it is  most easily recognisable (as indicated by high inter-annotator agreement; see  \citealt{2022-calabrese-plead}). We fine-tune Gemma-2-9B \cite{2024-riviere-gemma2} on PLEAD, and a sample of \gls{uplead} of comparable size, for up to 10 epochs. We then evaluate both models on the PLEAD test set and a disjoint sample of \gls{uplead}, also of comparable size.  Since \gls{uplead} is mostly balanced, we assume random subsamples will be approximately balanced too. Table~\ref{tab:slot-atomicity} reports production F1 scores (PF1; \citealt{2015-quirk-productionf1}) for all settings. 

Models achieve comparable performance when evaluated on a test set drawn from the same dataset used for training. Gemma trained on \gls{uplead} also performs well on the PLEAD test set, with only a ${\sim}{2}\%$ drop in performance.
Since \gls{uplead} was generated from a subset of expression clusters found in PLEAD, a few spans in the test set are unseen at training time.
Error analysis shows that the performance drop is mainly due to the model tagging more spans than found in the PLEAD goldstandard. Manual inspection of these false positives confirms that model  predictions often correspond to actual threats (e.g.,~\textit{kill}, \textit{axe}). However, in PLEAD  only one slot was annotated for posts containing \emph{multiple} harmful elements (e.g., dehumanizing comparisons and threats). This suggests the performance dip is mainly due to incomplete annotation and is, in practice, negligible.

%where no threats were annotated. However, this reflects a limitation in  PLEAD where for  posts with multiple forms of harmful content (e.g.,~dehumanising comparisons and threats), only one threat was annotated. 
%Manual inspection of false positives  revealed that the  model's span predictions did indeed correspond to threats (e.g.,~\textit{kill}, \textit{axe}). This suggests that this slight performance drop is largely an artifact of incomplete annotation and is, in practice, negligible.

\begin{table}[t]
\centering
\resizebox{0.56\linewidth}{!}{%
\begin{tabular}{llc} % Three columns, left-aligned
\toprule
Training & \multicolumn{1}{c}{Test} & \multicolumn{1}{c}{PF1} \\
\midrule
PLEAD  & PLEAD  & 73.82  \\
\gls{uplead}  & \gls{uplead}  & 76.05  \\
PLEAD  & \gls{uplead}  & 56.60  \\
\gls{uplead}  & PLEAD  & 73.02  \\
\bottomrule
\end{tabular}
}
\caption{Threatening speech classification (PF1\%) with Gemma-2-9B  trained/tested on  PLEAD and \gls{uplead}.}
\label{tab:slot-atomicity}
\end{table}

 In comparison, a model trained on PLEAD performs a lot worse when tested on \gls{uplead}  (there is a decrease of over $17\%$ between the two test sets). In this case, all spans are seen during training, and most errors on the \gls{uplead} test set are false negatives, where the model fails to recognise threats expressed in new contexts. For example, the threat \textit{``should be shot''} is correctly identified in \textit{``Dyke bitches are disgusting and should be shot,''} but not in \textit{``I'm curious to read that some people believe most Indigenous Australians are scum and should be shot.''} Our results suggest that Gemma does not treat slots as atomic concepts, even when other slots are not explicitly tagged, as the surrounding context influences the model’s ability to recognise a given slot. Training on a dataset in which correlation biases are mitigated appears to help Gemma better capture the atomicity of slot representations.

\subsection{\gls{uplead} Doesn't Degrade In-domain Performance}
\label{sec:impact:seen}

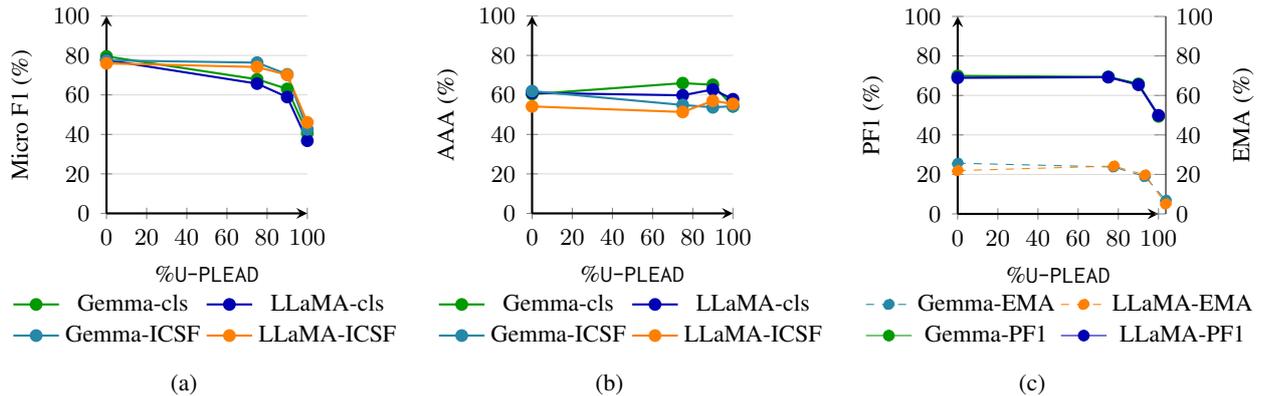
\begin{figure*}[t]
  \centering
  \begin{subfigure}[t]{0.3\textwidth}
    \centering
    \begin{tikzpicture}
     
     \begin{axis}[
    width=0.88\linewidth,
    height=4.2cm,
    axis lines=left,
    xlabel={\%\gls{uplead}},
    ylabel={Micro F1 (\%)},
    ymin=0, ymax=100,
    xmin=0, xmax=100,
    xtick={0, 20, ..., 100},
    ytick={0, 20, ..., 100},
    legend style={at={(0.5,-0.35)}, anchor=north, legend columns=2, draw=none, font=\small},
    grid=major,
    xmajorgrids=false,
    ymajorgrids=true,
    grid style={line width=.1pt, draw=gray!30},
    thick,
    mark size=2pt,
    every axis/.append style={
    font=\small  % or \footnotesize, \scriptsize, etc.
    }
]

% Gemma-classification (green)
\addplot[
    color=green!60!black,
    mark=*,
]
coordinates {
    (0,	79.50)
(75,	67.91)
(90,	63.03)
(100,	40.61)
};

% LLaMA-classification (dark blue)
\addplot[
    color=blue!70!black,
    mark=*,
]
coordinates {
    (0,	77.68)
(75,	65.71)
(90,	58.91)
(100,	36.78)
};

% Gemma-ICSF (light blue)
\addplot[
    color=cyan!60!black,
    mark=*,
]
coordinates {
    (0,	77.49)
(75,	76.34)
(90,	70.40)
(100,	42.53)
};

% LLaMA-ICSF (orange)
\addplot[
    color=orange,
    mark=*,
]
coordinates {
    (0,	75.96)
(75,	74.14)
(90,	70.11)
(100,	46.07)
};

\legend{
    Gemma-cls,
    LLaMA-cls,
    Gemma-ICSF,
    LLaMA-ICSF
}

\end{axis}
    \end{tikzpicture}
    \caption{\label{fig:iid-intent}}
  \end{subfigure}
  \hfill
  \begin{subfigure}[t]{0.3\textwidth}
    \centering
    \begin{tikzpicture}
    \begin{axis}[
    width=0.88\linewidth,
    height=4.2cm,
    axis lines=left,
    xlabel={\%\gls{uplead}},
    ylabel={AAA (\%)},
    ymin=0, ymax=100,
    xmin=0, xmax=100,
    xtick={0, 20, ..., 100},
    ytick={0, 20, ..., 100},
    legend style={at={(0.5,-0.35)}, anchor=north, legend columns=2, draw=none, font=\small},
    grid=major,
    xmajorgrids=false,
    ymajorgrids=true,
    grid style={line width=.1pt, draw=gray!30},
    thick,
    mark size=2pt,
    every axis/.append style={
    font=\small  % or \footnotesize, \scriptsize, etc.
    }
]

% Gemma-classification (green)
\addplot[
    color=green!60!black,
    mark=*,
]
coordinates {
    (0,	60.38)
(75,	65.95)
(90,	65.14)
(100,	54.17)
};

% LLaMA-classification (dark blue)
\addplot[
    color=blue!70!black,
    mark=*,
]
coordinates {
    (0,	61.02)
(75,	59.80)
(90,	62.74)
(100,	57.79)
};

% Gemma-ICSF (light blue)
\addplot[
    color=cyan!60!black,
    mark=*,
]
coordinates {
(0,	61.91)
(75,	54.90)
(90,	53.71)
(100,	54.30)
};

% LLaMA-ICSF (orange)
\addplot[
    color=orange,
    mark=*,
]
coordinates {
 (0,	54.16)
(75,	51.35)
(90,	57.00)
(100,	55.42)
};

\legend{
    Gemma-cls,
    LLaMA-cls,
    Gemma-ICSF,
    LLaMA-ICSF
}

\end{axis}
    \end{tikzpicture}
    \caption{\label{fig:iid-aaa}}
  \end{subfigure}
  \hfill
  \begin{subfigure}[t]{0.3\textwidth}
    \centering
    \begin{tikzpicture}
      % Primary axis (left) for classification models
\begin{axis}[
    width=0.88\linewidth,
    height=4.2cm,
    axis lines=left,
    xlabel={\%\gls{uplead}},
    ylabel={PF1 (\%)},
    ymin=0, ymax=100,
    xmin=0, xmax=100,
    xtick={0, 20, ..., 100},
    ytick={0, 20, ..., 100},
    grid=major,
    xmajorgrids=false,
    ymajorgrids=true,
    grid style={line width=.1pt, draw=gray!30},
    thick,
    mark size=2pt,
    every axis/.append style={font=\small}
]

% Gemma-PF1 (green)
\addplot[
    color=green!60!black,
    mark=*,
]
coordinates {
    (0, 69.93)
(75, 69.38)
(90, 65.85)
(100, 49.37)
};

% LLaMA-PF1 (dark blue)
\addplot[
    color=blue!70!black,
    mark=*,
]
coordinates {
    (0, 68.92)
(75, 69.24)
(90, 65.38)
(100, 49.92)
};

\end{axis}

% Secondary axis (right) for ICSF models
\begin{axis}[
    width=0.9\linewidth,
    height=4.2cm,
    axis y line*=right,
    axis x line=none,
    ylabel={EMA (\%)},
    ymin=0, ymax=100,
    xmin=0, xmax=100,
    ytick={0, 20, ..., 100},
    tick label style={font=\small},
    label style={font=\small},
    mark size=2pt,
    legend style={at={(0.5,-0.35)}, anchor=north, legend columns=2, draw=none, font=\small},
    % /tikz/every even column/.append style={column sep=0.5cm}
]

% Gemma-EMA (light blue)
\addplot[
    color=cyan!60!black,
    mark=*,
    dashed
]
coordinates {
    (0, 25.67)
(75, 23.75)
(90, 18.87)
(100, 6.90)
};

% LLaMA-EMA (orange)
\addplot[
    color=orange,
    mark=*,
    dashed
]
coordinates {
    (0, 21.93)
(75, 24.23)
(90, 19.73)
(100, 5.08)
};

% Combined legend
\addlegendimage{color=green!60!black, mark=*}          % Gemma-PF1
\addlegendimage{color=blue!70!black, mark=*}           % LLaMA-PF1
\addlegendimage{color=cyan!60!black, mark=*, dashed}   % Gemma-EMA
\addlegendimage{color=orange, mark=*, dashed}          % LLaMA-EMA
\legend{
    Gemma-EMA,
    LLaMA-EMA,
    Gemma-PF1,
    LLaMA-PF1,
}

\end{axis}
    \end{tikzpicture}
    \caption{\hspace{-1em}}
    \label{fig:iid-expl}
  \end{subfigure}
  \caption{Model performance on intent classification (cls and ICSF settings) using micro F1 (\subref{fig:iid-intent}) and AAA (\subref{fig:iid-aaa}). For ICSF models, we also show PF1 and EMA scores (\subref{fig:iid-expl}). Are results are reported on the PLEAD test set. }
  % and scores on intent classification on the PLEAD test set across different percentages of generated data for Gemma and LLaMA models, using classification (cls) and ICSF settings. For ICSF models, we also show PF1 and EMA scores (\subref{fig:iid-expl}).}
\end{figure*}
In our next experiment, we investigate whether training on \gls{uplead} mitigates correlation bias when evaluating on \emph{seen} combinations. There is no point generating data to handle generalisation if in-distribution performance is compromised. 

We focus on the full hate speech detection task (rather than classification of an individual slot) and  construct three training sets by combining instances from \gls{uplead}  and PLEAD at different ratios ($75\%$/$25\%$, $90\%$/$10\%$, and $100\%$/$0\%$) while keeping the total number of instances equal to that of the original PLEAD training set. We compare these configurations with a baseline trained solely on PLEAD (i.e., $0\%$/$100\%$). To ensure  our findings are not model-specific, we experiment with Gemma-2-9B and LLaMA-3.1-8B \cite{2024-grattafiori-llama3} considering two settings: intent classification (cls) and the more fine-grained task of intent classification \emph{and} slot filling (ICSF). %Models are fine-tuned for up to 25 and 15 epochs on the respective setting. 
For ICSF, we follow  \citet{2022-calabrese-plead} and  train models to only detect and fill slots, deterministically choosing the intent based on these. All models are evaluated on the testing partition of PLEAD. 

%the impact of mitigating correlation bias on the full hate speech detection task when evaluating on seen combinations --- under a scenario where the bias we aim to mitigate is still present at test time.

% To this end, we construct three training sets by combining instances from the \gls{uplead} dataset and PLEAD at different ratios ($75\%$/$25\%$, $90\%$/$10\%$, and $100\%$/$0\%$) while keeping the total number of instances equal to that of the original PLEAD training set. In the experiment, we compare these configurations with a baseline trained solely on PLEAD (i.e., $0\%$/$100\%$). To ensure that the findings from this experiment are not model-specific, we conduct experiments with both Gemma-2-9B and LLaMA-3.1-8B \cite{2024-grattafiori-llama3}. We consider two settings: a classification setting and an intent classification and slot filling (ICSF) setting. Models are fine-tuned for up to 25 and 15 epochs on the respective setting. Note that for the ICSF setting, we follow the approach introduced in \citet{2022-calabrese-plead}, training models only to detect and fill slots, while the intent is deterministically chosen based on the detected slots in the ontology.

We first observe that both Gemma and LLaMA achieve a Micro F1 score ${>}75\%$ on the intent classification task when trained exclusively on PLEAD, across all settings (cls and ICSF; see Figure~\ref{fig:iid-intent}). This is, to our knowledge, the highest score reported on the PLEAD benchmark to date---although with significantly larger models compared to \citet{2022-calabrese-plead}---indicating that we are starting from strong baselines. For both models, performance only slightly drops from classification to the harder ICSF task by~$2.1\%$ points, demonstrating that explainability no longer comes at the expense of performance.
Figure~\ref{fig:iid-intent} further illustrates that the effect of varying proportions of \gls{uplead} in training  depends on the setting rather than the model. ICSF models are robust to the replacement of~$75\%$ of training instances with \gls{uplead}, with a performance drop of less than~$2\%$ on intent classification. Although replacing $90\%$ of the data has a more noticeable impact, both models still achieve an F1 above~$70\%$. In contrast, training exclusively on \gls{uplead}  leads to a performance drop of nearly~$30\%$ for both models. %The behaviour of  classification models is markedly different. Performance drops for both models by  approximately~$12\%$ with the $75\%/25\%$ partition, and performance continues to decline as the proportion of \gls{uplead} increases. 
Classification models behave differently: performance drops~$12\%$ with the $75\%$/$25\%$ partition and continues to decline as more \gls{uplead} data is introduced.

%Both patterns are consistent with our expectations. 
\gls{uplead} is by design constructed so that text spans do not correlate with intent labels, thereby providing limited learning signal for models that treat the post holistically. The gradual introduction of \gls{uplead} instances counters the tendency classification models have to rely on superficial cues or shortcuts, as further evidenced in Figure~\ref{fig:iid-intent} by the higher AAA scores (with Gemma improving by $5.57\%$ at $75\%$ \gls{uplead} and LLaMA by $1.72\%$ at $90\%$ \gls{uplead}). ICSF models are trained solely on slot labels, and although \gls{uplead} removes correlations between slot and intent labels, spans are still consistently annotated with the same  label (e.g., \textit{``kill''} is always labelled as \texttt{SL:ThreateningSpeech}, regardless of context), providing models with a more robust learning signal. 

Training exclusively on \gls{uplead} prevents the model from learning useful information about the target distribution, such as the typical structure of parse trees observed in the dataset, leading to a drop in performance (Appendix~\ref{sec:model-output-examples}). \gls{uplead} seems less beneficial to ICSF, according to  AAA scores. This may be due to the models' improved ability to detect slots in new, not necessarily smoothly connected, contexts, which in turn makes them more sensitive to the adversarial nature of the AAA evaluation procedure. For the ICSF models we also evaluate production F1 (PF1) and tree exact match (EMA) in Figure~\ref{fig:iid-expl}. Both models achieve comparable high performance when fine-tuned on PLEAD: $69.93\%$ and $25.67\%$ for Gemma and $68.92\%$ and $21.93\%$ for LLaMa, respectively. We observe similar trends to intent classification (Figure~\ref{fig:iid-intent}) when increasing the proportion of training \gls{uplead} instances. 

In sum, up to $75\%$ of the training set can be replaced with \gls{uplead} instances without  significantly affecting the performance of  ICSF models on classification and parsing metrics.

\subsection{\gls{uplead}  Improves Generalisation}

We now investigate whether training on \gls{uplead} supports compositional generalisation on the full hate speech detection task. In these experiments, we evaluate models on the \texttt{TARGET} benchmark and expect them to handle novel combinations of (possibly unseen) targets and expressions.
Recall that by design the \texttt{TARGET} benchmark contains span combinations that do not appear in \gls{uplead}. 
%training data. This is not necessarily the case for models trained on PLEAD, which may have already encountered some of the test-time combinations during training.

We use the same training sets from the previous experiment (Section~\ref{sec:impact:seen})  combining different proportions of  \gls{uplead}  and and PLEAD.
%at different ratios ($75\%$/$25\%$, $90\%$/$10\%$, and $100\%$/$0\%$) while keeping the total number of instances equal to that of the original PLEAD training set. We compare these configurations with a baseline trained solely on PLEAD (i.e., $0\%$/$100\%$). To ensure  our findings are not model-specific, we conduct experiments with both Gemma-2-9B and LLaMA-3.1-8B \cite{2024-grattafiori-llama3} considering two settings: intent classification and the more fine-grained tasks of intent classification \emph{and} slot filling (ICSF). %Models are fine-tuned for up to 25 and 15 epochs on the respective setting. 
%For ICSF, we follow  \citet{2022-calabrese-plead} and  train models to only detect and fill slots, deterministically chosing the intent based on these slots.% in the ontology.
% We evaluate Gemma and LLaMA fine-tuned with $75\%$ \gls{uplead} instances on the compositional generalisation benchmark introduced earlier, considering both the classification and ICSF settings, and compare their performance with that of models trained exclusively on PLEAD. For completeness, results for the $90\%$/$10\%$ and $100\%$/$0\%$ training sets are reported in Appendix~\ref{sec:gen-tests}. 
%
Table~\ref{tab:intent-gen-tests} reports  Micro F1 on intent classification (computed as  the geometric mean of the scores obtained across the eight generalisation tests in \texttt{TARGET}) for models trained on the $75\%$/$25\%$ and $0\%$/$100\%$ training sets. For results on $90\%$/$10\%$ and $100\%$/$0\%$ see Appendix~\ref{sec:additional-results}. Our experiments show that neither the model architecture nor the setting has a significant effect on performance, whereas training on \gls{uplead}  (combined with some  fraction of PLEAD) consistently leads to higher classification scores. Overall performance remains low, with the best model achieving a Micro F1 of $50.34\%$ which is substantially lower than the estimated annotation accuracy of~$76.25$\% (Section~\ref{sec:seen}). This is due to the challenging nature of the task rather than imperfections in the automatically generated benchmark.
\begin{table}[t]
\centering
\resizebox{0.82\linewidth}{!}{%
\begin{tabular}{l c r c}
\toprule
Model & Setting & \multicolumn{1}{c}{\%\gls{uplead}} & Micro F1  \\
\midrule
Gemma & Cls & 0\%   & 46.18 \\
Gemma & Cls & 75\%  & \textbf{49.22} \\
% Gemma & Class. & 90\%  & \textbf{51.53} \\
% Gemma & Class. & 100\% & 41.92 \\
\midrule
Gemma & ICSF           & 0\%   & 46.94 \\
Gemma & ICSF           & 75\%  & \textbf{50.34} \\
% Gemma & ICSF           & 90\%  & \textbf{50.51} \\
% Gemma & ICSF           & 100\% & 46.75 \\
\midrule
LLaMA & Cls & 0\%   & 46.84 \\
LLaMA & Cls & 75\%  & \textbf{49.53} \\
% LLaMA & Class. & 90\%  & 47.14 \\
% LLaMA & Class. & 100\% & 43.03 \\
\midrule
LLaMA & ICSF           & 0\%   & 46.51 \\
LLaMA & ICSF           & 75\%  & \textbf{49.03} \\
% LLaMA & ICSF           & 90\%  & 47.97 \\
% LLaMA & ICSF           & 100\% & 47.23 \\
\bottomrule
\end{tabular}
}
\caption{Model performance (Micro F1) on \texttt{TARGET}  using different proportions of \gls{uplead} for training. Results are reported for intent classification (Cls) alone and in combination with slot filling (ICSF).}
%classification and ICSF settings across different settings and proportions of generated data.}
\label{tab:intent-gen-tests}
\end{table}
\begin{table}[t]
\centering
\resizebox{0.72\linewidth}{!}{%
\begin{tabular}{l r c c}
\toprule
Model & \multicolumn{1}{c}{\%\gls{uplead}} & PF1  & EMA \\
\midrule
Gemma & 0\%   & 22.95 & 0.50 \\
Gemma & 75\%  & \textbf{44.98} & \textbf{5.91} \\
% Gemma & 90\%  & \textbf{46.53} & \textbf{6.06} \\
% Gemma & 100\% & 44.72 & 5.45 \\
\midrule
LLaMA & 0\%   & 22.84 &  0.51 \\
LLaMA & 75\%  & \textbf{44.56} & \textbf{5.51}  \\
% LLaMA & 90\%  & 45.05 &  5.55 \\
% LLaMA & 100\% & \textbf{45.21} &  \textbf{5.61} \\
\bottomrule
\end{tabular}
}
\caption{Model performance  on intent classification and slot filling task (ICSF) using production F1  (PF1) and exact match (EMA) metrics. Results are reported on the 
\texttt{TARGET} benchmark using different proportions of \gls{uplead} for training.}
\label{tab:pf1-gen-tests}
\end{table}

Table~\ref{tab:pf1-gen-tests} summarises model performance on the intent classification \emph{and} slot filling task using production F1 (PF1) and tree exact match (EMA) as evaluation metrics. 
We observe the  most substantial impact of \gls{uplead} on  the PF1 metric, where performance improves by over~$21\%$;  the  EMA score also  increases by more than $5\%$. These gains suggest that the model produces more accurate slot annotations, leading to higher-quality explanations for its predictions. Improving explanation quality is particularly important for content moderation. \citet{2024-calabrese-explainability} show that structured explanations can make professional moderators faster, contingent on these being reliable. % when deciding whether to rely on model assistance.

Evaluation across individual generalisation tests follows a similar pattern (see Appendix~\ref{sec:additional-results}). However, two outliers emerge: in Tests \hyperlink{desc:t3}{3} and~\hyperlink{desc:t3b}{3b}, Gemma and LLaMA achieve lower Micro F1 scores in the ICSF setting when trained with \gls{uplead} instances compared to training on PLEAD. In these tests, all examples belong to  the non-hateful intent class. Training on \gls{uplead}  enhances the model’s ability to recognise slot spans in unfamiliar contexts, which in turn increases its tendency to tag slots overall. Since intent labels are deterministically assigned based on  predicted slots, %a greater number of predicted slots raises the likelihood of triggering a policy rule, leading to \emph{incorrect} intent predictions in cases that should be classified as non-hateful. 
this raises the chance of incorrectly triggering a policy rule in non-hateful cases.
We still observe substantial improvements on parsing metrics in both tests, and the classification setting continues to show gains in Micro F1, confirming the broader benefits of \gls{uplead} even in these edge cases.

% \section{Discussion}
% \input{08_discussion}
\section{Conclusions}
We empirically demonstrate that ICSF models struggle with compositional generalisation in the context of hate speech.
We propose a theoretically motivated experimental setting that ensures full  coverage of each expression’s behaviour within the training set. We argue this  is ideal for learning the underlying dynamics of hate speech by explicitly removing spurious correlations between expressions, labels, and targets.

To facilitate this analysis, we introduce \gls{uplead}, a (mostly) balanced synthetic dataset, alongside \acrshort{genbench}, the first benchmark for evaluating compositional generalisation in hate speech. Although classification models do not produce trees, the function they are expected to learn still implicitly involves identifying targets and hateful expressions, making \acrshort{genbench} a relevant generalisation test for them as well. Our experiments reveal that augmenting real training data with a portion of \gls{uplead} improves generalisation while maintaining state-of-the-art performance on in-domain test sets. However, training exclusively on synthetic data leads to a decline in performance which points to the difficulty of improving model generalisation on test sets whose distribution diverges from training while maintaining performance on in-domain data. 

Our findings show that ICSF models achieve performance comparable to standard classification models, indicating that explainability no longer comes at the cost of accuracy. As generative models become more common for classification, the computational overhead of ICSF is also less, leaving little justification for treating hate speech detection as a black-box problem.

\section*{Ethical Statement}
This project is motivated by the need for more robust and fair methods to detect hateful language online reliably and transparently. Our work involved the generation of hate speech content for the explicit purpose of improving the fairness and robustness of hate speech detection models. To support reproducibility and facilitate future research, we will release the code, the finetuned models, and our datasets (\gls{uplead}, \acrshort{genbench}) under the CC BY-NC-SA 4.0 License. However, due to the sensitive nature of the content, access to these will be granted upon request and subject to a declaration of intent, in order to prevent misuse\footnote{For access to datasets, code or models, please contact a.calabrese@ed.ac.uk.}.

The \gls{uplead} and \acrshort{genbench} datasets were produced using large language models, including those developed by OpenAI. While this required prompting to generate harmful language, we did not need to develop sophisticated jailbreak techniques. We present this as evidence of existing vulnerabilities in the safety mechanisms of LLMs, and do not endorse the use of such techniques outside of controlled, research-driven settings.

We observed instances of stereotypical associations in the generated generalisation tests, such as the use of personal names linked to specific ethnicities. These unintended biases further highlight the limitations of current generation models and underscore the importance of using the benchmark exclusively for evaluation.

In this work, we follow a prescriptive paradigm \cite{2022-rottger-paradigm}, and all resources have been created under the assumption that each post has a single true label, determined by the policy.

\section*{Limitations}
Our work is currently limited to English, due to the lack of structured annotation resources in other languages. However, our grammar-based generation procedure would apply to any hate speech data annotated with intents and slots. 

While our generalisation benchmark is entirely synthetic, this design choice is necessary to precisely control slot and expression combinations. Collecting a sufficiently large and diverse set of real-world posts that reflect these controlled configurations is not currently feasible. As the benchmark is automatically generated, some noise and limitations in linguistic naturalness are to be expected. However, given the current capabilities of large language models --- and with an estimated $76.25\%$ accuracy in intent labels annotation --- we find the quality of the benchmark sufficient for evaluating compositional generalisation in a realistic setting, especially considering that model performance remains well below this accuracy ceiling. Importantly, our evaluation is not limited to synthetic data: we also report results on the original PLEAD test set.

Finally, while the training data has not been manually validated, the results of the experiment in Section~\ref{sec:exp-slot} demonstrate that the dataset is informative enough to improve generalisation, indicating that it provides a useful learning signal. 
\section*{Acknowledgements}
We would like to thank Matthias Lindemann and Filip Smola for their valuable feedback and helpful comments.
This work was supported in part by Huawei and the UKRI Centre for Doctoral Training in Natural Language Processing, funded by the UKRI (grant EP/S022481/1) and the University of Edinburgh, School of Informatics. 
Calabrese gratefully acknowledges the support of the Turing AI Fellowship: Neural Conversational Information Seeking Assistant (grant EP/V025708/1). Ross gratefully acknowledges the support of the Royal Society of Edinburgh, RSE Saltire Facilitation Network Award, award no. 1901. Lapata gratefully acknowledges the support of the UK Engineering
and Physical Sciences Research Council (grant
EP/L016427/1).

\begin{figure}[h!]
    \centering
    \begin{subfigure}[b]{0.15\textwidth}
        \centering
        \includegraphics[height=6.2mm]{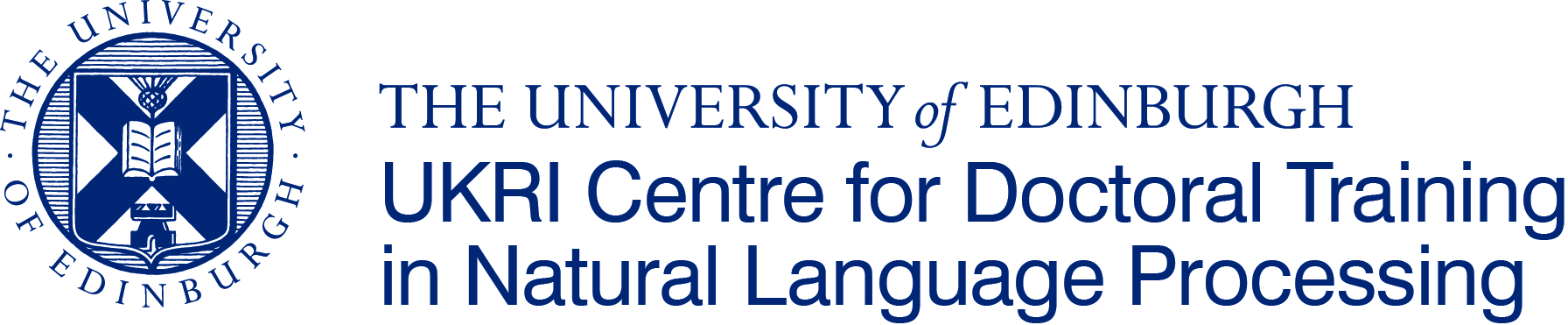}
    \end{subfigure}
    \hfill
    \begin{subfigure}[b]{0.08\textwidth}
        \centering
        \includegraphics[height=5.7mm]{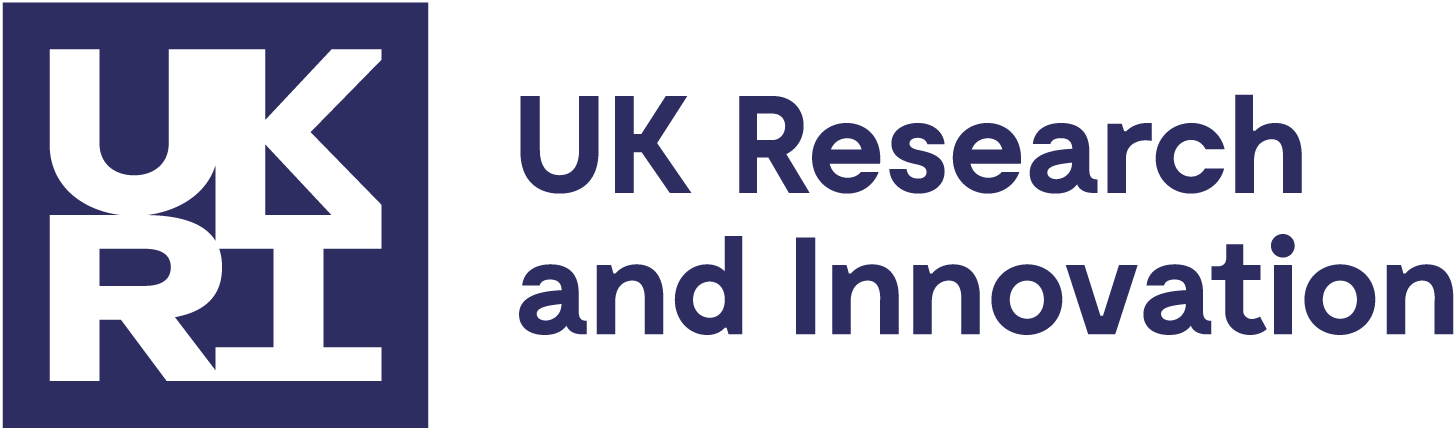}
    \end{subfigure}
    \hfill
    \begin{subfigure}[b]{0.15\textwidth}
        \centering
        \includegraphics[height=6.2mm]{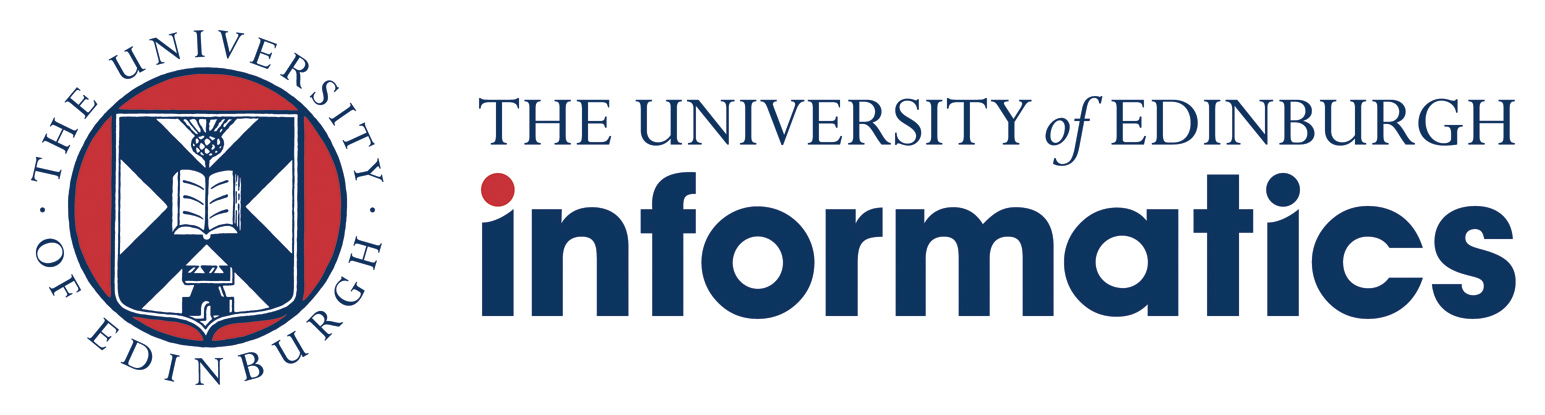}
    \end{subfigure}
\end{figure}

% \section*{Acknowledgments}

\bibliography{custom}

\appendix
\section{The PLEAD Dataset}
\label{sec:plead}
\begin{figure}[h]
    \centering
    \noindent
    \begin{forest}
      for tree={
        edge={->},
        parent anchor=south,
        child anchor=north,
        align=center,
        l sep=6pt,
        s sep=8pt,
        font=\small
      }
      [\texttt{IN:NotHateful}
        [\texttt{SL:Target}
          [\textit{"Artists"}]
        ]
        [\texttt{SL:DehumanisingComparison}
          [\textit{"are parasites"}]
        ]
      ]
    \end{forest}
    \caption{Example of slot and intent annotations for the non-hateful post \textit{``Artists are parasites''}.}
    \label{fig:plead-example}
\end{figure}
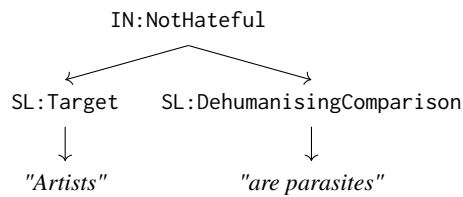

The PLEAD dataset\footnote{\url{https://huggingface.co/datasets/agostina3/PLEAD}} was built by enriching a subset of hateful and non-hateful posts from the LFTW dataset \cite{2021-vidgen-lftw} with span-level annotations. This enabled the framing of hate speech detection as an instance of intent classification and slot filling (ICSF), where slots represent elements such as the ``target'' or ``protected characteristic'', while intents correspond to policy rules (see Figure~\ref{fig:plead-example}).
The trees follow a decoupled representation \cite{2020-Aghajanyan-conversational}, where spans may appear multiple times and in any order relative to the original post. Not all tokens in the post need to be included in the tree, and spans may connect non-adjacent tokens.

PLEAD contains 3,535 English posts and is split into training, validation and test set in an 80/10/10 ratio, with each split preserving the original intent distribution: non-hateful ($25\%$), dehumanisation ($25\%$), threatening ($17\%$), derogation ($28\%$), and support of hate crimes ($5\%$). Each post includes three annotated parse trees, resulting in $\sim{}11,000$ training instances. The best-performing model in the literature is a BART-based architecture with a two-step generation approach \cite{2022-calabrese-plead}, which achieved a micro F1 score of $57.17\%$ on intent classification and a PF1 score of $52.96\%$.

Our use of the dataset complies with the terms of the CC BY-NC-SA 4.0 License.

\section{Tree Generation}
\label{sec:tree_generation}
In this section we provide more details on how we generated the collection of trees for \gls{uplead} while satisfying the constraints described in Section~\ref{sec:bias-free}.

\paragraph{Equivalence Among Terminal Symbols} As described in Section~\ref{sec:bias-free}, we apply hierarchical clustering to define equivalence relationships among terminal symbols in the hate speech grammar $G$, with the exception of protected targets, which are grouped based on the already available target annotations in PLEAD (e.g., ``woman'', ``she'', and ``her'' are grouped under ``women''). The clustering threshold of 0.5 was selected empirically based on cluster quality across a range of values. For slots with a limited number of spans (e.g., hate entities and negative stance), we first augment the data by prompting GPT-3.5-Turbo with existing spans, asking it to complete the list.

\paragraph{Constraint Enforcement} For the generation of \gls{uplead} we select $40$ clusters of protected targets and hate entities, and $20$ clusters for each remaining slot. When more clusters are available than needed, we select  in decreasing  size order. Constraint \hyperlink{desc:c5}{C5} requires that each protected and non-protected target and each hate entity occur with each dehumanising comparison, threat, negative opinion, and expression of support for hate crimes with same frequency. As an example, consider the cluster of protected targets $t_1$. To satisfy the constraint, $t_1$ must occur with any cluster $i$ of dehumanising comparisons $d_i$, threats $t_{h_i}$, negative opinions $n_i$ and expressions of support $s_i$ the same number of times~$n$. For the first three slots, this is straightforward:
\[
\mathop{\forall}\limits_{i} \quad \left|\{ \text{tree} \mid t_1 \in \text{tree} \land d_i \in \text{tree} \} \right| = n
\]
\[
\mathop{\forall}\limits_{i} \quad \left|\{ \text{tree} \mid t_1 \in \text{tree} \land t_{h_i} \in \text{tree} \} \right| = n
\]
\[
\mathop{\forall}\limits_{i} \quad \left|\{ \text{tree} \mid t_1 \in \text{tree} \land n_i \in \text{tree} \} \right| = n
\]
Let's now consider the intents corresponding to these trees. A tree containing the protected target $t_1$ and the dehumanising comparison $d_i$ can have only two possible intents: \texttt{IN:DehumanisingComparison} or \texttt{IN:NotHateful} --- the latter applies when the tree also includes a negative stance expression. Symmetrically, the trees containing $t_1$ and $t_{h_i}$ or $n_i$ can only have two possible intents: \texttt{IN:NotHateful} and \texttt{IN:ThreateningSpeech} or \texttt{IN:Derogation}, respectively. Suppose we define $n$ trees for each cluster of support expressions $s_i$ to satisfy \hyperlink{desc:c5}{C5} as described above:
\[
\mathop{\forall}\limits_{i} \quad \left|\{ \text{tree} \mid t_1 \in \text{tree} \land s_i \in \text{tree} \} \right| = n
\]
Since expressions of support can yield hateful intents only when associated to hate entities, and not protected targets, the $n$ trees so defined would all have one possible intent only: \texttt{IN:NotHateful}. Constraint \hyperlink{desc:c1}{C1} requires $t_1$ to occur with the same frequency across all classes. Because $t_1$ would occur in $n$ non-hateful trees, we would be forced to assign all $n$ trees containing $t_1$ and $d_i$ to the dehumanisation class, without introducing any negative stance expression. Symmetrically, all trees combining $t_1$ with $t_{h_i}$ or $n_i$ would belong to the corresponding hateful classes. The result of these choices would be a dataset with no instances of counter speech\footnote{Posts that may quote hate speech but where the author explicitly rejects or disapproves of the harmful message.}, a phenomenon that we want models to be able to recognise, and an over-representation of non-hateful instances with the same target-support structure, introducing a bias the model should not learn.

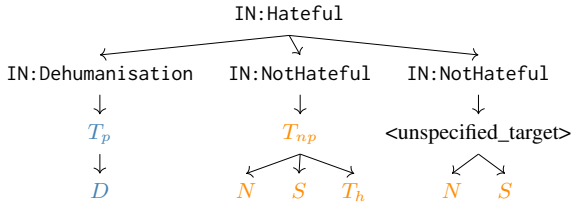
\begin{figure}[t]
    \centering
    \noindent
    \small{Main Tree: \textcolor{slot1color}{$T_p D$}  Injected Slots: \textcolor{slot2color}{$N N T_{np} S S T_h$}\\}
    \vspace{1em}
    \resizebox{\linewidth}{!}{%
   \begin{forest}
      for tree={
        edge={->},
        parent anchor=south,
        child anchor=north,
        align=center,
        l sep=5pt,
        s sep=8pt,
        font=\small,
      }
      [\texttt{IN:Hateful}
      [\texttt{IN:Dehumanisation}
        [\textcolor{slot1color}{$T_p$}
          [\textcolor{slot1color}{$D$}]
        ]
    ]
    [\texttt{IN:NotHateful}
        [\textcolor{slot2color}{$T_{np}$}
            [\textcolor{slot2color}{$N$}]
            [\textcolor{slot2color}{$S$}]
            [\textcolor{slot2color}{$T_h$}]
        ]
    ]
    [\texttt{IN:NotHateful}
        [<unspecified\_target>
            [\textcolor{slot2color}{$N$}]
            [\textcolor{slot2color}{$S$}]
        ]
    ]
    ]
    \end{forest}
    }
\caption{Example of how the main-tree and injected slots associated with an instance are organised into a tree.}
\label{fig:tree-structure}
\end{figure}

Therefore, we need to find another way to combine $t_1$ with $s_i$ $n$ times. We introduce the concepts of \textit{main tree} and \textit{injected slots}. The main tree of an example $e$ is the set of slots that determines the intent of $e$, and can include up to 4 slots. The injected slots of $e$ are slots that occur in the tree, but do not affect the final intent classification (see Figure~\ref{fig:tree-structure}). Note that the grammar is designed to generate a flat set of slot annotations for each post, without any inherent hierarchical structure or distinction between types of slots. The characterisation of one set of slots as the ``main tree'' and others as ``injections'' is imposed as a post-processing step, added to satisfy the constraints and facilitate the subsequent translation of trees into posts. For reference, all instances in PLEAD would have the annotated slots in the main tree, and 0 injected slots. With this characterisation, we can satisfy \hyperlink{desc:c5}{C5} and associate $t_1$ with $s_i$ $n$ times as follows:
\[
\mathop{\forall}\limits_{i} \left|\{ \text{tree} \mid t_1 \in \text{main(tree)} \land s_i \in \text{main(tree)} \} \right| = m
\]
\[
\mathop{\forall}\limits_{i} \, \left|\{ \text{tree} \mid t_1 \in \text{main(tree)} \land s_i \in \text{inj(tree)} \} \right| = o
\]
\[
n = m + o
\]
where $main(x)$ is a function that returns the set of slots in the main tree of $x$, and $inj(x)$ is a function that returns the set of injected slots in $x$.

The distinction between main and injected slots is also central to satisfying constraints \hyperlink{desc:c2}{C2}, \hyperlink{desc:c3}{C3}, and \hyperlink{desc:c4}{C4}, as it allows us to inject, for example, a threat, a hate entity, or a negative stance expression into an instance labeled as dehumanisation without altering its class. While satisfying the constraints requires many instances to include injected slots, we also aim to preserve simpler cases that contain only main-tree slots. To balance this, we enforce that ${\sim}70\%$ of the trees in \gls{uplead} contain no injections, while the other half cover all required injections --- possibly resulting in the generation of some large and complex trees. Table~\ref{tab:trees-distribution} provides an overview of the tree structures in \gls{uplead} and the number of instances corresponding to each structure. For instance, we generate $12{,}800$ trees with the structure $T_pD$ in the main tree and no injected slots. Here, we slightly abuse notation by using $T_p$ to indicate $T$ restricted to protected targets. Since we select~$40$ clusters of protected targets and $20$ clusters of dehumanising comparisons, this results in $16$ trees for each cluster pair $\langle{}t_i, d_i\rangle$ (i.e.,xv$n=16$). When clusters contain a sufficient number of spans, we sample without replacement to diversify the instances in the dataset; otherwise, sampling is done with replacement.

\begin{table*}[]
\centering
\resizebox{0.7\linewidth}{!}{%
\begin{tabular}{llr llr}
\toprule
\textbf{Main Tree} & \multicolumn{1}{c}{\textbf{Injected Slots}} & \multicolumn{1}{c}{\textbf{N}} &
\multicolumn{1}{c}{\textbf{Main Tree}} & \multicolumn{1}{c}{\textbf{Injected Slots}} & \multicolumn{1}{c}{\textbf{N}} \\
\midrule
$S$ &  & 39200 & $D$ &  & 29600 \\
$T_h$ &  & 29600 & $N$ &  & 29600 \\
$C$ &  & 29600 & $T_p C$ &  & 25600 \\
$T_p D$ &  & 12800 & $T_p T_h$ &  & 12800 \\
$T_p S$ &  & 12800 & $T_p N$ &  & 12800 \\
$T_p N$ & $D T_{np} S S T_h T_h$ & 3786 & $T_p D$ & $N N T_{np} S S T_h$ & 3737 \\
$T_p T_h$ & $D N N T_{np} S S$ & 3736 & $T_p N$ & $D D T_{np} S S T_h$ & 3718 \\
$T_p D$ & $N T_{np} S S T_h T_h$ & 3699 & $T_p T_h$ & $D D N T_{np} S S$ & 3686 \\
$T_p D$ & $N N N_s T_{np} S S T_h$ & 3333 & $T_p T_h$ & $D N N N_s T_{np} S S$ & 3297 \\
$T_p T_h$ & $D D N N_s T_{np} S S$ & 3295 & $T_p D$ & $N N_s T_{np} S S T_h T_h$ & 3264 \\
$T_p N$ & $D N_s T_{np} S S T_h T_h$ & 3254 & $T_p N$ & $D D N_s T_{np} S S T_h$ & 3241 \\
$E S$ &  & 3200 & $T_p D N_s$ &  & 3200 \\
$T_p N_s T_h$ &  & 3200 & $T_p N_s S$ &  & 3200 \\
$T_p N N_s$ &  & 3200 & $T_{np} D$ &  & 3200 \\
$T_{np} T_h$ &  & 3200 & $T_{np} S$ &  & 3200 \\
$T_{np} N$ &  & 3200 & $T_{np} C$ &  & 3200 \\
$T_p S$ & $D N N_s T_h T_h$ & 3005 & $T_p S$ & $D N N_s T_{np} T_h T_h$ & 2995 \\
$E D$ &  & 2400 & $E T_h$ &  & 2400 \\
$E N_s S$ &  & 2400 & $E N$ &  & 2400 \\
$T_p S$ & $D N T_h T_h$ & 2289 & $T_p D$ & $E N N S S T_h$ & 2275 \\
$T_p T_h$ & $D E N N S S$ & 2273 & $T_p N$ & $D D E S S T_h$ & 2237 \\
$T_p D$ & $E N S S T_h T_h$ & 2200 & $T_p T_h$ & $D D E N S S$ & 2189 \\
$T_p N$ & $D E S S T_h T_h$ & 2184 & $T_p S$ & $D N T_{np} T_h T_h$ & 2167 \\
$T_p D$ & $E N N_s S S T_h T_h$ & 2012 & $T_p N$ & $D D E N_s S S T_h$ & 1997 \\
$T_p T_h$ & $D D E N N_s S S$ & 1984 & $T_p N$ & $D E N_s S S T_h T_h$ & 1983 \\
$T_p T_h$ & $D E N N N_s S S$ & 1940 & $T_p D$ & $E N N N_s S S T_h$ & 1880 \\
$T_p S$ & $D D N N_s T_{np} T_h$ & 1847 & $T_p S$ & $D D N N_s T_h$ & 1767 \\
$T_p S$ & $D D N T_{np} T_h$ & 1359 & $T_p S$ & $D D N T_h$ & 1343 \\
$T_p S$ & $D N N N_s T_h$ & 1034 & $T_p S$ & $D N N N_s T_{np} T_h$ & 986 \\
$E S$ & $D N T_{np}$ & 822 & $E S$ & $N T_{np} T_h$ & 817 \\
$T_p S$ & $D N N T_{np} T_h$ & 810 & $E S$ & $D N$ & 797 \\
$T_p S$ & $D N N T_h$ & 765 & $E S$ & $N T_h$ & 764 \\
$T_p S$ & $D N N T_h T_h$ & 755 & $T_p S$ & $D N N T_{np} T_h T_h$ & 723 \\
$E S$ & $N_s T_h$ & 645 & $T_p S$ & $D D N N N_s T_{np}$ & 609 \\
$E S$ & $N_s T_{np} T_h$ & 591 & $E S$ & $D N_s$ & 590 \\
$E S$ & $D N_s T_{np}$ & 574 & $T_p N$ & $D N_s T_{np} S T_h T_h$ & 560 \\
$T_p S$ & $D D N N N_s$ & 557 & $T_p D$ & $N N_s T_{np} S T_h T_h$ & 546 \\
$T_p T_h$ & $D N N N_s T_{np} S$ & 529 & $T_p T_h$ & $D D N N_s T_{np} S$ & 528 \\
$T_p N$ & $D D N_s T_{np} S T_h$ & 515 & $T_p D$ & $N N N_s T_{np} S T_h$ & 502 \\
$T_p T_h$ & $D D N N_s N_s T_{np} S$ & 486 & $T_p N$ & $D D N_s N_s T_{np} S T_h$ & 482 \\
$T_p S$ & $D D N N T_{np} T_h$ & 477 & $T_p S$ & $D D N N T_h$ & 473 \\
$T_p D$ & $N N N_s N_s T_{np} S T_h$ & 470 & $T_p D$ & $N N_s N_s T_{np} S T_h T_h$ & 449 \\
$T_p S$ & $D D N N T_{np}$ & 446 & $T_p N$ & $D N_s N_s T_{np} S T_h T_h$ & 444 \\
$T_p T_h$ & $D N N N_s N_s T_{np} S$ & 443 & $T_p S$ & $D D N N$ & 421 \\
$T_p T_h$ & $D D E N N_s S$ & 331 & $T_p T_h$ & $D E N N N_s S$ & 328 \\
$T_p D$ & $E N N N_s S T_h$ & 323 & $T_p D$ & $E N N_s S T_h T_h$ & 318 \\
$T_p D$ & $E N N_s N_s S T_h T_h$ & 312 & $T_p N$ & $D D E N_s N_s S T_h$ & 306 \\
$T_p N$ & $D D E N_s S T_h$ & 304 & $T_p T_h$ & $D D E N N_s N_s S$ & 301 \\
$T_p N$ & $D E N_s S T_h T_h$ & 296 & $T_p N$ & $D E N_s N_s S T_h T_h$ & 293 \\
$T_p D$ & $E N N N_s N_s S T_h$ & 280 & $T_p T_h$ & $D E N N N_s N_s S$ & 254 \\
$T_p S$ & $D N N N T_{np} T_h$ & 237 & $T_p S$ & $D N N N T_h$ & 234 \\
$E S$ & $D T_{np}$ & 212 & $E S$ & $D$ & 205 \\
$E S$ & $T_h$ & 199 & $E S$ & $T_{np} T_h$ & 184 \\
$T_p S$ & $D D N N N$ & 157 & $T_p S$ & $D D N N N T_{np}$ & 144 \\
\bottomrule
\end{tabular}
}
\caption{Distribution of tree structures in \gls{uplead}. Each group shows a main tree structure, the injected slots (if any), and the number of matching instances. We use $T_p$ and $T_{np}$ to distinguish between protected and non-protected targets.}
\label{tab:trees-distribution}
\end{table*}

% Explain how injections are handled in Appendix C?
\paragraph{Tree Post-Processing} In the post-processing stage, we split the flat set of slots associated with each instance into a main tree and injected slots, assigning them a hierarchical structure. Since the final intent label of an instance is, by definition, determined by the main tree, the injected slots must be organised into subtrees in such a way that they only yield non-hateful intents (see Figure~\ref{fig:tree-structure}). We satisfy this constraint by never allocating $D$, $T_h$ and $N$ to subtrees with protected targets. Symmetrically, we never allocate $S$ to subtrees with hate entities. To prevent individual subtrees from becoming too large, we allocate at most three occurrences of any of the slots $D$, $T_h$, $N$, and $S$ per subtree. Non-protected targets and hate entities, if present, are distributed across the available subtrees. When the total number of non-protected targets and hate entities in an instance is smaller than the number of subtrees (excluding the main tree), we assign the special target token \texttt{<unspecified\_target>} to the remaining subtrees. As a result, these subtrees may contain, for example, a threat or a derogatory opinion without a clearly identified target, reflecting instances where the intended target is left implicit in the post. Splitting the slots into subtrees (i.e., opinions) will also help us with the translation of the trees into posts, as it allows to break down the instructions into smaller prompts.

While trees in PLEAD are limited to three levels --- corresponding to intents, slots, and text spans --- we modify this structure by rooting the slots $D$, $T_h$, $N$, $S$ and $N_s$ under the target or hate entity they are directed at. For protected targets, we additionally root the protected characteristic slot beneath its corresponding target. This hierarchical structure better captures instances where multiple expressions are directed at the same target. Finally, we add an additional binary intent layer at the top of the tree to summarise the intents of all subtrees within an instance. This layer is necessary to aggregate subtree-level intent labels without requiring the selection of a specific hateful intent, particularly in cases where multiple subtrees express different types of hateful content (e.g., threatening and derogatory). In fact, while \gls{uplead} is designed such that only the main tree carries a hateful intent, a model trained under this framework may still produce outputs with multiple hateful subtrees. 
\section{Post Generation}
\label{sec:post-generation}
We convert the trees generated by our grammar into posts by first generating a draft of the posts using Vicuna-30B-uncensored\footnote{\url{https://huggingface.co/cognitivecomputations/Wizard-Vicuna-30B-Uncensored}} in a few-shot setting. The instruction specifies which spans must occur in the post (possibly verbatim), and the role each span must play, together with a few in-context examples. The examples are selected from PLEAD to align with the structure of the tree for which we are generating a post (when possible). For instance, to generate a prompt associated to a tree with structure $TX$ ($X \in \{D, T_h, N\})$ we would prompt Vicuna with the following instruction:
\begin{tcolorbox}
\small 
\textbf{USER:} \textit{Write a post where ``Polish people'' is target of the opinion ``are so nasty'' because it belongs to the group ``Polish''. ``Polish people'' shouldn't occur at the beginning of the post.}\\
\textbf{ASSISTANT:} \textit{Truly, I have been to Poland and there’s nothing to see there. Polish people are so nasty as well.}\\
\\
\textbf{USER:} \textit{Write a post where ``women'' is target of the opinion ``should be shot'' because it belongs to the group ``women''. ``women'' shouldn't occur at the beginning of the post.}\\
\textbf{ASSISTANT:} \textit{Women should be shot? Agreed!}\\
\\
\textbf{USER:} \textit{Write a post where ``a breastfeeding woman'' is target of the opinion ``can do it at home'' because it belongs to the group ``she''. ``a breastfeeding woman'' shouldn't occur at the beginning of the post.}\\
\textbf{ASSISTANT:} \textit{I don’t want to have to be around a breastfeeding woman, if she wants to do that she can do it at home}\\
\\
\textbf{USER:} $\langle{}input\_tree\rangle\\$
\textbf{ASSISTANT:}
\end{tcolorbox}
We request the span corresponding to the target to not occur at the beginning of the post to prevent creating a dataset with strong positional biases.

To improve fluency and ensure the spans are included in the posts, we refine them using GPT-3.5-Turbo. We simulate a two-step conversation by providing our original instructions and Vicuna-generated output as messages in the history, prompting GPT to improve ``its'' previous response by simply repeating the instructions:
\begin{tcolorbox}
\small 
\textbf{USER:}  $\langle{}Vicuna\_instruction\rangle$\\
\textbf{ASSISTANT:}  $\langle{}Vicuna\_output\rangle$\\
\textbf{USER:}  $\langle{}Vicuna\_instruction\rangle$\\
\textbf{ASSISTANT:}
\end{tcolorbox}

If any of the requested spans cannot be found in the generated post we perform additional generation rounds. We instruct GPT to explicitly tag where each span appears in the post as follows:
\begin{tcolorbox}
\small 
\textbf{USER:} \textit{Post: $\langle{}GPT\_output\rangle$\\
Copy and integrate all the following phrases in the post verbatim:
\begin{itemize}
    \item <span\_0> span </span\_0>
    \item <span\_1> span </span\_1>
    \item <span\_n> span </span\_n>
\end{itemize}
When inserting the spans in the post, include the corresponding tags to mark start and end. The result should be a single coherent post.}\\
\textbf{ASSISTANT:}
\end{tcolorbox}
If the GPT output indicates the presence of a span that does not exactly match the requested one, we apply our clustering algorithm and retain the span only if it belongs to the same cluster as the requested span. 
\section{Generalisation to Unseen Combinations}
\label{sec:experiments-details}
\subsection{Evaluation Metrics}
\label{sec:eval-metrics}
In this section, we define the evaluation metrics used in this study. For all metrics, higher values indicate better performance.

\paragraph{Intent Classification} We evaluate model performance on the intent classification task, which is equivalent to traditional black-box classification in a multi-class setting, using Micro F1. This metric treats each prediction equally, regardless of its class, and therefore helps account for class imbalance in the dataset. For models trained on \gls{uplead}, which may generate multiple trees per instance, we define the final predicted intent to be \texttt{IN:NotHateful} if and only if \textit{all} generated trees have this intent. For hateful instances, the prediction is considered correct if \textit{any} of the generated trees contains the same hateful intent as the gold annotation.

To assess robustness under more challenging conditions, we also report performance using the AAA (Adversarial Attacks against Abuse) metric \cite{2021-calabrese-aaa}. AAA evaluates intent classification in a binary setting, where all hateful intents are grouped under a single label (\texttt{IN:Hateful}), and \texttt{IN:NotHateful} remains unchanged. It penalizes models that rely on shallow lexical features by adversarially modifying the test set based on patterns seen during training. The final AAA score is computed as the geometric mean of the F1 scores across 4 adversarial scenarios. To date, models evaluated on AAA have performed no better than random.

\paragraph{ICSF} We evaluate model performance on the full ICSF task by measuring production F1 (PF1) and tree exact match accuracy (EMA). PF1 is computed by representing each parse tree as a set of production rules and calculating the F1 score over these sets \cite{2015-quirk-productionf1}.  In our setup, the productions are extracted at the token level: for example, tagging the span \textit{``are parasites''} as \texttt{SL:DehumanisingComparison} yields two productions: one linking the slot to \textit{``are''} and one to \textit{``parasites''}. 

EMA measures the percentage of predictions in which the entire set of productions exactly matches the reference annotation for an instance. The comparison is set-based, meaning that all slots and their corresponding spans must be correct, but the order in which they are produced does not affect the score. 

Since both PLEAD and \acrshort{genbench} provide only a single annotated tree per instance, while models trained on \gls{uplead} can generate multiple candidate trees, we evaluate ICSF metrics only on the output tree that is closest to the reference.

\subsection{Parameters}
\label{sec:parameters}
We fine-tuned Gemma-2-9B\footnote{\url{https://huggingface.co/google/gemma-2-9b}} and LLaMA-3.1-8B\footnote{\url{https://huggingface.co/meta-llama/Llama-3.1-8B}} on a NVIDIA H100-80GB GPU with a per-device batch size of 2 and gradient accumulation over 4 steps. We used a learning rate of $2\times10^{-4}$ with a linear learning rate scheduler and 5 warmup steps. We used the AdamW optimizer with 8-bit precision and applied a weight decay of 0.01. The classification and ICSF models were fine-tuned for up to 25 and 15 epochs, respectively. Fine-tuning Gemma for 1 epoch required $\sim{}12$ minutes for classification and $\sim{}53$ minutes for ICSF. Statistics for LLaMA are $7$ minutes/epoch for classification and $\sim{33}$ minutes/epoch for ICSF.

For inference we sample only one output with nucleus sampling probability of 0.95 and temperaure set to 0.8.

\subsection{Model Output Examples}
\label{sec:model-output-examples}
This section contains examples of outputs generated by Gemma when trained using different proportions of \gls{uplead}.

\begin{figure*}[]
\centering
\parbox{\textwidth}{\centering
    \footnotesize{Post: \textit{``Has this happened to anyone? I just can't put aside racial evolution, it just doesn't make sense to. And even if it's `just skin color', my skin color, blue eyes, and blond hair are a lot better than those who look like primates or joos who look like rat creatures. Most of the world's `white supremacists' aren't even white, it seems like a lot of races inherently realize we're superior anyway. Like a filipino girl asked me once `how does it feel to be god's favorite'. Is there any way to hypothetically untake the redpill? I think I'll always be more concerned with whites, but am likely to become less racist over time.''}}
}
\\
\vspace{0.5cm}
\begin{subfigure}[t]{0.53\textwidth}
\centering
\parbox{\textwidth}{\centering
    \footnotesize{Ground truth in PLEAD}
}
\vspace{0.2cm}
\resizebox{\textwidth}{!}{%
    \begin{forest}
    for tree={grow'=south, child anchor=north, parent anchor=south, edge={->}, align=center, l sep=5pt, font=\small}
    [\texttt{IN:Hateful}
    [\texttt{IN:Dehumanisation}
        [\texttt{SL:Target}\\\textit{joos}
        [\texttt{SL:ProtectedCharacteristic}\\\textit{joos}]
        [\texttt{SL:DehumanisingComparison}\\\textit{like primates or joos who}\\\textit{look like rat creatures}]
        ]
    ]
    ]
    \end{forest}
}
\end{subfigure}
\\
\vspace{0.3cm}
\begin{subfigure}[t]{0.32\textwidth}
\centering
\parbox{0.5\textwidth}{\centering
    \footnotesize{0\% \gls{uplead}}
}
\vspace{0.3cm}
\parbox{0.5\textwidth}{\centering
    \footnotesize{ICSF}
}
\resizebox{\textwidth}{!}{%
    \begin{forest}
    for tree={grow'=south, child anchor=north, parent anchor=south, edge={->}, align=center, l sep=5pt, font=\small}
    [\texttt{IN:Hateful}
    [\texttt{IN:Dehumanisation}
        [\texttt{SL:Target}\\\textit{joos}
        [\texttt{SL:Protected}\\\texttt{Characteristic}\\\textit{joos}]
        [\texttt{SL:Dehumanising}\\\texttt{Comparison}\\\textit{who look like}\\\textit{rat creatures}]
        ]
    ]
    ]
    \end{forest}
}
\end{subfigure}
\hfill
\begin{subfigure}[t]{0.32\textwidth}
\centering
\parbox{0.5\textwidth}{\centering
    \footnotesize{75\% \gls{uplead}}
}
\vspace{0.3cm}
\parbox{0.5\textwidth}{\centering
    \footnotesize{ICSF}
}
\resizebox{\textwidth}{!}{%
    \begin{forest}
    for tree={grow'=south, child anchor=north, parent anchor=south, edge={->}, align=center, l sep=5pt, font=\small}
    [\texttt{IN:Hateful}
        [\texttt{IN:Dehumanisation}
        [\texttt{SL:Target}\\\textit{joos}
        [\texttt{SL:Protected}\\\texttt{Characteristic}\\\textit{joos}]
        [\texttt{SL:Dehumanising}\\\texttt{Comparison}\\\textit{rat creatures}]
        ]
    ]
    ]
    \end{forest}
}
\end{subfigure}
\hfill
\begin{subfigure}[t]{0.32\textwidth}
\centering
\parbox{0.5\textwidth}{\centering
    \footnotesize{90\% \gls{uplead}}
}
\vspace{0.3cm}
\parbox{0.5\textwidth}{\centering
    \footnotesize{ICSF}
}
\resizebox{\textwidth}{!}{%
    \begin{forest}
    for tree={grow'=south, child anchor=north, parent anchor=south, edge={->}, align=center, l sep=5pt, font=\small}
    [\texttt{IN:Hateful}
        [\texttt{IN:Dehumanisation}
        [\texttt{SL:Target}\\\textit{those who look like rat creatures}
        [\texttt{SL:Protected}\\\texttt{Characteristic}\\\textit{joos}]
        [\texttt{SL:Dehumanising}\\\texttt{Comparison}\\\textit{rat creatures}]
        ]
    ]
    ]
    \end{forest}
}
\end{subfigure}
\\
\vspace{0.3cm}
\begin{subfigure}[t]{\textwidth}
\centering
\parbox{0.5\textwidth}{\centering
    \footnotesize{100\% \gls{uplead}}
}
\vspace{0.3cm}
\parbox{0.5\textwidth}{\centering
    \footnotesize{ICSF}
}
\resizebox{\textwidth}{!}{%
    \begin{forest}
    for tree={grow'=south, child anchor=north, parent anchor=south, edge={->}, align=center, l sep=5pt, font=\small}
    [\texttt{IN:Hateful}
        [\texttt{IN:NotHateful}
            [\texttt{SL:Target}\\<unspecified\_target>
            [\texttt{SL:Threatening}\\\texttt{Speech}\\\textit{untake the redpill}]
            [\texttt{SL:Support}\\\texttt{HateCrimes}\\\textit{i think i'll always be}\\\textit{more concerned with whites}\\\textit{but am likely to become less}\\\textit{racist over time}]
            ]
        ]
        [\texttt{IN:NotHateful}
            [\texttt{SL:HateEntity}\\\textit{most of the world's  white supremacists}
            [\texttt{SL:Dehumanising}\\\texttt{Comparison}\\\textit{look like primates}]
            [\texttt{SL:Dehumanising}\\\texttt{Comparison}\\\textit{joos who look like}\\\textit{rat creatures}]
            ]
        ]
         [\texttt{IN:Derogation}
            [\texttt{SL:Target}\\\textit{a filipino girl}
            [\texttt{SL:Protected}\\\texttt{Characteristic}\\\textit{filipino}]
            [\texttt{SL:Negative}\\\texttt{Opinion}\\\textit{how does it feel to}\\\textit{be god's favorite}]
            ]
        ]
    ]
    \end{forest}
}
\end{subfigure}
\\
\vspace{0.5cm}
\parbox{0.2\textwidth}{\centering
    \footnotesize{0\% \gls{uplead}}\\
    \vspace{0.3cm}
    \footnotesize{Classification\\\texttt{Dehumanisation}}
}
\hfill
\parbox{0.2\textwidth}{\centering
    \footnotesize{75\% \gls{uplead}}\\
    \vspace{0.3cm}
    \footnotesize{Classification\\\texttt{Dehumanisation}}
}
\hfill
\parbox{0.2\textwidth}{\centering
    \footnotesize{90\% \gls{uplead}}\\
    \vspace{0.3cm}
    \footnotesize{Classification\\\texttt{Dehumanisation}}
}
\hfill
\parbox{0.2\textwidth}{\centering
    \footnotesize{100\% \gls{uplead}}\\
    \vspace{0.3cm}
    \footnotesize{Classification\\\texttt{Dehumanisation}}
}
\caption{Example of a hateful post from PLEAD and how Gemma's predictions change depending on the proportion of \gls{uplead} used for training, under both the ICSF and classification settings.}
\label{fig:model-output-plead-ex1}
\end{figure*}

\paragraph{Experiments on PLEAD} The experiments in Section~\ref{sec:impact:seen} show that up to 75\% of the PLEAD training set can be replaced with \gls{uplead} instances without significantly affecting the in-domain performance of ICSF models, both in terms of classification and parsing accuracy. Figure~\ref{fig:model-output-plead-ex1} illustrates this by comparing the output of the Gemma model on a hateful PLEAD post under different training settings. When 75\% of the training data is replaced, the model’s prediction remains largely consistent with the original, demonstrating robust retention of the target structure. However, as the proportion of synthetic data increases to 90\%, the model begins to show noticeable changes, and when trained exclusively on \gls{uplead}, its output diverges substantially. In this case, the model fails to reflect key characteristics of the original PLEAD data --- such as the typical structure of parse trees --- and instead produces outputs with many more slot annotations. While each individual slot label may appear reasonable, the way they are combined is incorrect, leading to misinterpretation of the overall intent. The prediction of the classification model on the other hand remained correct across all the training settings.

\begin{figure*}[]
\centering
\parbox{\textwidth}{\centering
    \footnotesize{Post: \textit{``The Wang family , who are Taiwanese , have been predating on is my pastime for years. It's my pastime now to seek revenge on them.''}}
}
\\
\vspace{0.5cm}
\begin{subfigure}[t]{0.4\textwidth}
\centering
\parbox{\textwidth}{\centering
    \footnotesize{Ground truth in \acrshort{genbench}}
}
\vspace{0.2cm}
\resizebox{\textwidth}{!}{%
    \begin{forest}
    for tree={grow'=south, child anchor=north, parent anchor=south, edge={->}, align=center, l sep=5pt, font=\small}
    [\texttt{IN:Threatening}
        [\texttt{SL:Target}\\\textit{The Wang Family}
        [\texttt{SL:ProtectedCharacteristic}\\\textit{Taiwanese}]
        [\texttt{SL:ThreateningSpeech}\\\textit{predating on is my pastime}]
        ]
    ]
    \end{forest}
}
\end{subfigure}
\\
\vspace{0.3cm}
\begin{subfigure}[t]{0.24\textwidth}
\centering
\parbox{0.5\textwidth}{\centering
    \footnotesize{0\% \gls{uplead}}
}
\vspace{0.3cm}
\parbox{0.5\textwidth}{\centering
    \footnotesize{ICSF}
}
\resizebox{\textwidth}{!}{%
    \begin{forest}
    for tree={grow'=south, child anchor=north, parent anchor=south, edge={->}, align=center, l sep=5pt, font=\small}
    [\texttt{IN:Threatening}
        [\texttt{SL:Target}\\\textit{The Wang Family}
        [\texttt{SL:Protected}\\\texttt{Characteristic}\\\textit{Wang}]
        [\texttt{SL:Threatening}\\\texttt{Speech}\\\textit{who are}\\\textit{Taiwanese}]
        ]
    ]
    \end{forest}
}
\end{subfigure}
\hfill
\begin{subfigure}[t]{0.24\textwidth}
\centering
\parbox{0.5\textwidth}{\centering
    \footnotesize{75\% \gls{uplead}}
}
\vspace{0.3cm}
\parbox{0.5\textwidth}{\centering
    \footnotesize{ICSF}
}
\resizebox{\textwidth}{!}{%
    \begin{forest}
    for tree={grow'=south, child anchor=north, parent anchor=south, edge={->}, align=center, l sep=5pt, font=\small}
    [\texttt{IN:Threatening}
        [\texttt{SL:Target}\\\textit{The Wang Family}
        [\texttt{SL:Protected}\\\texttt{Characteristic}\\\textit{Taiwanese}]
        [\texttt{SL:Threatening}\\\texttt{Speech}\\\textit{now to}\\\textit{seek revenge}]
        ]
    ]
    \end{forest}
}
\end{subfigure}
\hfill
\begin{subfigure}[t]{0.24\textwidth}
\centering
\parbox{0.5\textwidth}{\centering
    \footnotesize{90\% \gls{uplead}}
}
\vspace{0.3cm}
\parbox{0.5\textwidth}{\centering
    \footnotesize{ICSF}
}
\resizebox{\textwidth}{!}{%
    \begin{forest}
    for tree={grow'=south, child anchor=north, parent anchor=south, edge={->}, align=center, l sep=5pt, font=\small}
    [\texttt{IN:Threatening}
        [\texttt{SL:Target}\\\textit{The Wang Family}
        [\texttt{SL:Protected}\\\texttt{Characteristic}\\\textit{Taiwanese}]
        [\texttt{SL:Threatening}\\\texttt{Speech}\\\textit{predating on }\\\textit{is my pastime}]
        ]
    ]
    \end{forest}
}
\end{subfigure}
\hfill
\begin{subfigure}[t]{0.24\textwidth}
\centering
\parbox{0.5\textwidth}{\centering
    \footnotesize{100\% \gls{uplead}}
}
\vspace{0.3cm}
\parbox{0.5\textwidth}{\centering
    \footnotesize{ICSF}
}
\resizebox{\textwidth}{!}{%
    \begin{forest}
    for tree={grow'=south, child anchor=north, parent anchor=south, edge={->}, align=center, l sep=5pt, font=\small}
    [\texttt{IN:Threatening}
        [\texttt{SL:Target}\\\textit{The Wang Family}
        [\texttt{SL:Protected}\\\texttt{Characteristic}\\\textit{Taiwanese}]
        [\texttt{SL:Threatening}\\\texttt{Speech}\\\textit{predating on}\\\textit{is my pastime}]
        ]
    ]
    \end{forest}
}
\end{subfigure}
\\
\vspace{0.5cm}
\parbox{0.2\textwidth}{\centering
    \footnotesize{Classification\\\texttt{Threatening}}
}
\hfill
\parbox{0.2\textwidth}{\centering
    \footnotesize{Classification\\\texttt{Threatening}}
}
\hfill
\parbox{0.2\textwidth}{\centering
    \footnotesize{Classification\\\texttt{Dehumanisation}}
}
\hfill
\parbox{0.2\textwidth}{\centering
    \footnotesize{Classification\\\texttt{Not-Hateful}}
}
\caption{Example of a hateful post from Test 1 in \acrshort{genbench} and how Gemma's predictions change depending on the proportion of \gls{uplead} used for training, under both the ICSF and classification settings.}
\label{fig:model-output-gen-ex1}
\end{figure*}

\begin{figure*}[]
\centering
\parbox{\textwidth}{\centering
    \footnotesize{Post: \textit{``Despotic despotic leaders are so underrated. it's definitely ok for to hate anyone who disagrees with that opinion.''}}
}
\\
\vspace{0.5cm}
\begin{subfigure}[t]{0.4\textwidth}
\centering
\parbox{\textwidth}{\centering
    \footnotesize{Ground truth in \acrshort{genbench}}
}
\vspace{0.2cm}
\resizebox{0.5\textwidth}{!}{%
    \begin{forest}
    for tree={grow'=south, child anchor=north, parent anchor=south, edge={->}, align=center, l sep=5pt, font=\small}
    [\texttt{IN:ProHateCrimes}
        [\texttt{SL:HateEntity}\\\textit{Despotic}
        [\texttt{SL:SupportHateCrimes}\\\textit{it's definitely ok for to hate}]
        ]
    ]
    \end{forest}
}
\end{subfigure}
\\
\vspace{0.3cm}
\begin{subfigure}[t]{0.24\textwidth}
\centering
\parbox{0.5\textwidth}{\centering
    \footnotesize{0\% \gls{uplead}}
}
\vspace{0.3cm}
\parbox{0.5\textwidth}{\centering
    \footnotesize{ICSF}
}
\resizebox{\textwidth}{!}{%
    \begin{forest}
    for tree={grow'=south, child anchor=north, parent anchor=south, edge={->}, align=center, l sep=5pt, font=\small}
    [\texttt{IN:NotHateful}
        [\texttt{SL:Target}\\\textit{Despotic despotic leaders}
        [\texttt{SL:NegativeOpinion}\\\textit{are so underrated}]
        ]
    ]
    \end{forest}
}
\end{subfigure}
\hfill
\begin{subfigure}[t]{0.24\textwidth}
\centering
\parbox{0.5\textwidth}{\centering
    \footnotesize{75\% \gls{uplead}}
}
\vspace{0.3cm}
\parbox{0.5\textwidth}{\centering
    \footnotesize{ICSF}
}
\resizebox{\textwidth}{!}{%
    \begin{forest}
    for tree={grow'=south, child anchor=north, parent anchor=south, edge={->}, align=center, l sep=5pt, font=\small}
    [\texttt{IN:NotHateful}
        [\texttt{SL:HateEntity}\\\textit{The Wang Family}
        [\texttt{SL:NegativeOpinion}\\\textit{it's definitely ok for to hate}]
        ]
    ]
    \end{forest}
}
\end{subfigure}
\hfill
\begin{subfigure}[t]{0.24\textwidth}
\centering
\parbox{0.5\textwidth}{\centering
    \footnotesize{90\% \gls{uplead}}
}
\vspace{0.3cm}
\parbox{0.5\textwidth}{\centering
    \footnotesize{ICSF}
}
\resizebox{\textwidth}{!}{%
    \begin{forest}
    for tree={grow'=south, child anchor=north, parent anchor=south, edge={->}, align=center, l sep=5pt, font=\small}
    [\texttt{IN:ProHateCrimes}
        [\texttt{SL:HateEntity}\\\textit{Despotic}
        [\texttt{SL:SupportHateCrimes}\\\textit{it's definitely ok for to hate}]
        ]
    ]
    \end{forest}
}
\end{subfigure}
\hfill
\begin{subfigure}[t]{0.24\textwidth}
\centering
\parbox{0.5\textwidth}{\centering
    \footnotesize{100\% \gls{uplead}}
}
\vspace{0.3cm}
\parbox{0.5\textwidth}{\centering
    \footnotesize{ICSF}
}
\resizebox{\textwidth}{!}{%
    \begin{forest}
    for tree={grow'=south, child anchor=north, parent anchor=south, edge={->}, align=center, l sep=5pt, font=\small}
    [\texttt{IN:ProHateCrimes}
        [\texttt{SL:HateEntity}\\\textit{Despotic}
        [\texttt{SL:SupportHateCrimes}\\\textit{it's definitely ok for to hate}]
        ]
    ]
    \end{forest}
}
\end{subfigure}
\\
\vspace{0.5cm}
\parbox{0.2\textwidth}{\centering
    \footnotesize{Classification\\\texttt{Not-Hateful}}
}
\hfill
\parbox{0.2\textwidth}{\centering
    \footnotesize{Classification\\\texttt{Not-Hateful}}
}
\hfill
\parbox{0.2\textwidth}{\centering
    \footnotesize{Classification\\\texttt{Not-Hateful}}
}
\hfill
\parbox{0.2\textwidth}{\centering
    \footnotesize{Classification\\\texttt{Not-Hateful}}
}
\caption{Example of a hateful post from Test $1_b$ in \acrshort{genbench} and how Gemma's predictions change depending on the proportion of \gls{uplead} used for training, under both the ICSF and classification settings.}
\label{fig:model-output-gen-ex2}
\end{figure*}

\paragraph{Experiments on \acrshort{genbench}} When evaluating on the \acrshort{genbench} benchmark we observe a different trend, with Gemma's outputs getting closer to the annotated tree as more instances from \gls{uplead} are included in the training set. For instance, Figure~\ref{fig:model-output-gen-ex1} shows a case where the intent prediction remains correct across all ICSF training settings, while the accuracy of the output tree improves gradually. In contrast, for the same example, the classification setting does not benefit from the inclusion of \gls{uplead} data: Gemma's prediction shifts from the correct hateful intent to an incorrect, yet still hateful one, and ultimately to a non-hateful intent.

Figure~\ref{fig:model-output-gen-ex1} shows an example where improvements in the accuracy of the generated tree lead to a shift from an incorrect to a correct intent prediction. In contrast, the classification model’s prediction remains consistently incorrect across all training settings.

\subsection{Additional results on \acrshort{genbench}}
\label{sec:additional-results}

\begin{table}[]
\centering
\resizebox{0.85\linewidth}{!}{%
\begin{tabular}{l c r c}
\toprule
Model & Setting & \multicolumn{1}{c}{\%\gls{uplead}} & Micro F1  \\
\midrule
% Gemma & Cls & 0\%   & 46.18 \\
% Gemma & Cls & 75\%  & 49.22 \\
Gemma & Cls & 90\%  & 51.53 \\
Gemma & Cls & 100\% & 41.92 \\
\midrule
% Gemma & ICSF           & 0\%   & 46.94 \\
% Gemma & ICSF           & 75\%  & 50.34 \\
Gemma & ICSF           & 90\%  & 50.51 \\
Gemma & ICSF           & 100\% & 46.75 \\
\midrule
% LLaMA & Cls & 0\%   & 46.84 \\
% LLaMA & Cls & 75\%  & \textbf{49.53} \\
LLaMA & Cls & 90\%  & 47.14 \\
LLaMA & Cls & 100\% & 43.03 \\
\midrule
% LLaMA & ICSF           & 0\%   & 46.51 \\
% LLaMA & ICSF           & 75\%  & \textbf{49.03} \\
LLaMA & ICSF           & 90\%  & 47.97 \\
LLaMA & ICSF           & 100\% & 47.23 \\
\bottomrule
\end{tabular}
}
\caption{Model performance (Micro F1) on \texttt{TARGET} using different proportions of \gls{uplead} for training. Results are reported for intent classification (Cls) alone and in combination with slot filling (ICSF).}
%classification and ICSF settings across different settings and proportions of generated data.}
\label{tab:intent-gen-tests-v2}
\end{table}
\begin{table}[]
\centering
\resizebox{0.8\linewidth}{!}{%
\begin{tabular}{l r c c}
\toprule
Model & \multicolumn{1}{c}{\%\gls{uplead}} & PF1  & EMA \\
\midrule
% Gemma & 0\%   & 22.95 & 0.50 \\
% Gemma & 75\%  & 44.98 & 5.91 \\
Gemma & 90\%  & 46.53 & 6.06 \\
Gemma & 100\% & 44.72 & 5.45 \\
\midrule
% LLaMA & 0\%   & 22.84 &  0.51 \\
% LLaMA & 75\%  & 44.56 & 5.51  \\
LLaMA & 90\%  & 45.05 &  5.55 \\
LLaMA & 100\% & 45.21 &  5.61 \\
\bottomrule
\end{tabular}
}
\caption{Model performance  on intent classification and slot filling task (ICSF) using production F1  (PF1) and exact match (EMA) metrics. Results are reported on the 
\texttt{TARGET} benchmark using different proportions of \gls{uplead} for training.}
\label{tab:pf1-gen-tests-v2}
\end{table}

\begin{table*}[t]
\centering
\resizebox{0.65\linewidth}{!}{%
\begin{tabularx}{0.73\linewidth}{l c r c c c}
\toprule
Model & Setting & \% Gen. Data & Micro F1 (\%) & PF1 (\%) & EMA (\%) \\
\midrule
Gemma & Cls & 0\%   & 50.62 & --- & --- \\
Gemma & Cls & 75\%  & \textbf{53.40} & --- & --- \\
Gemma & Cls & 90\%  & 52.53 & --- & --- \\
Gemma & Cls & 100\% & 50.30 & --- & --- \\
\midrule
Gemma & ICSF & 0\%   & 56.25 & 40.96 & 1.39 \\
Gemma & ICSF & 75\%  & 60.14 & 53.44 & 6.70 \\
Gemma & ICSF & 90\%  & \textbf{60.84} & \textbf{54.81} & \textbf{7.76} \\
Gemma & ICSF & 100\% & 55.13 & 53.24 & 6.10 \\
\midrule
LLaMA & Cls & 0\%   & 50.87 & --- & --- \\
LLaMA & Cls & 75\%  & 51.88 & --- & --- \\
LLaMA & Cls & 90\%  & \textbf{53.42} & --- & --- \\
LLaMA & Cls & 100\% & 51.31 & --- & --- \\
\midrule
LLaMA & ICSF           & 0\%   & 53.84 & 38.48 & 1.09 \\
LLaMA & ICSF           & 75\%  & \textbf{56.03} & 52.51 & \textbf{6.37} \\
LLaMA & ICSF           & 90\%  & 55.78 & \textbf{53.66} & 6.10 \\
LLaMA & ICSF           & 100\% & 53.42 & 52.27 & 6.08 \\
\bottomrule
\end{tabularx}
}
\caption{Micro F1, PF1 and EMA scores on \acrshort{genbench}'s Test 1 for Gemma-2-9B and LLaMA-3.1-8B across different settings and proportions of generated data.}
\label{tab:gen-test-1}
\end{table*}

\begin{table*}[t]
\centering
\resizebox{0.65\linewidth}{!}{%
\begin{tabularx}{0.73\linewidth}{l c r c c c}
\toprule
Model & Setting & \% Gen. Data & Micro F1 (\%) & PF1 (\%) & EMA (\%) \\
\midrule
Gemma & Cls & 0\%   & 56.40 & --- & --- \\
Gemma & Cls & 75\%  & \textbf{56.89} & --- & --- \\
Gemma & Cls & 90\%  & 55.41 & --- & --- \\
Gemma & Cls & 100\% & 53.86 & --- & --- \\
\midrule
Gemma & ICSF & 0\%   & \textbf{60.41} & 43.79 & 1.92 \\
Gemma & ICSF & 75\%  &  59.70 & 53.48 & 9.61 \\
Gemma & ICSF & 90\%  &  59.29 & \textbf{53.76} & \textbf{9.70} \\
Gemma & ICSF & 100\% &  54.80 & 52.65 & 8.98 \\
\midrule
LLaMA & Cls & 0\%   & 56.02 & --- & --- \\
LLaMA & Cls & 75\%  &  52.96 & --- & --- \\
LLaMA & Cls & 90\%  &  \textbf{56.50} & --- & --- \\
LLaMA & Cls & 100\% &  52.79 & --- & --- \\
\midrule
LLaMA & ICSF           & 0\%   & \textbf{59.34} & 41.73 & 1.36 \\ 
LLaMA & ICSF           & 75\%  &  58.42 & \textbf{52.82} & \textbf{9.19} \\ 
LLaMA & ICSF           & 90\%  &  55.36 & 52.80 & 8.73 \\ 
LLaMA & ICSF           & 100\% &  53.98 & 51.24 & 8.20 \\ 
\bottomrule
\end{tabularx}
}
\caption{Micro F1, PF1 and EMA scores on \acrshort{genbench}'s Test 2 for Gemma-2-9B and LLaMA-3.1-8B across different settings and proportions of generated data.}
\label{tab:gen-test-2}
\end{table*}

\begin{table*}[]
\centering
\resizebox{0.65\linewidth}{!}{%
\begin{tabularx}{0.73\linewidth}{l c r c c c}
\toprule
Model & Setting & \% Gen. Data & Micro F1 (\%) & PF1 (\%) & EMA (\%) \\
\midrule
Gemma & Cls & 0\%   & 59.37 & --- & --- \\
Gemma & Cls & 75\%  & 70.59 & --- & --- \\
Gemma & Cls & 90\%  & \textbf{77.22} & --- & --- \\
Gemma & Cls & 100\% & 71.02 & --- & --- \\
\midrule
Gemma & ICSF & 0\%   & \textbf{75.81} & 40.93 & 0.14 \\
Gemma & ICSF & 75\%  & 63.18  & 53.55  & \textbf{3.40} \\
Gemma & ICSF & 90\%  & 64.71  & \textbf{54.04}  & 3.35 \\
Gemma & ICSF & 100\% & 59.10  & 51.86  & 2.94 \\
\midrule
LLaMA & Cls & 0\%   & 70.91 & --- & --- \\
LLaMA & Cls & 75\%  &  72.22 & --- & --- \\
LLaMA & Cls & 90\%  &  \textbf{74.04} & --- & --- \\
LLaMA & Cls & 100\% &  67.78 & --- & --- \\
\midrule
LLaMA & ICSF           & 0\%   & \textbf{68.87} & 39.03 & 0.24 \\ 
LLaMA & ICSF           & 75\%  &  63.46 & 52.63 & 2.18 \\ 
LLaMA & ICSF           & 90\%  &  55.89 & 51.90 & 2.97 \\ 
LLaMA & ICSF           & 100\% &  58.15 & \textbf{53.73} & \textbf{3.32} \\ 
\bottomrule
\end{tabularx}
}
\caption{Micro F1, PF1 and EMA scores on \acrshort{genbench}'s Test 3 for Gemma-2-9B and LLaMA-3.1-8B across different settings and proportions of generated data.}
\label{tab:gen-test-3}
\end{table*}

\begin{table*}[t]
\centering
\resizebox{0.65\linewidth}{!}{%
\begin{tabularx}{0.73\linewidth}{l c r c c c}
\toprule
Model & Setting & \% Gen. Data & Micro F1 (\%) & PF1 (\%) & EMA (\%) \\
\midrule
Gemma & Cls & 0\%   & 50.62 & --- & --- \\
Gemma & Cls & 75\%  &  52.50 & --- & --- \\
Gemma & Cls & 90\%  &  \textbf{52.95} & --- & --- \\
Gemma & Cls & 100\% &  51.08 & --- & --- \\
\midrule
Gemma & ICSF & 0\%   & 55.41 & 38.95 & 0.96 \\
Gemma & ICSF & 75\%  &  57.44 & 49.35 & 4.89 \\
Gemma & ICSF & 90\%  &  \textbf{57.67} & \textbf{51.12} & \textbf{5.04} \\
Gemma & ICSF & 100\% &  53.99 & 49.72 & 4.31 \\
\midrule
LLaMA & Cls & 0\%   & 50.72 & --- & --- \\
LLaMA & Cls & 75\%  & 51.36 & --- & --- \\
LLaMA & Cls & 90\%  & \textbf{53.23} & --- & --- \\
LLaMA & Cls & 100\% & 51.76 & --- & --- \\
\midrule
LLaMA & ICSF           & 0\%   & 51.76 & 35.99 & 0.79 \\
LLaMA & ICSF           & 75\%  &  \textbf{54.17} & 48.07 & \textbf{4.66} \\
LLaMA & ICSF           & 90\%  &  52.98 & \textbf{49.20} & 4.01 \\
LLaMA & ICSF           & 100\% &  50.87 & 48.63 & 4.41 \\
\bottomrule
\end{tabularx}
}
\caption{Micro F1, PF1 and EMA scores on \acrshort{genbench}'s Test 4 for Gemma-2-9B and LLaMA-3.1-8B across different settings and proportions of generated data.}
\label{tab:gen-test-4}
\end{table*}

\begin{table*}[t]
\centering
\resizebox{0.65\linewidth}{!}{%
\begin{tabularx}{0.73\linewidth}{l c r c c c}
\toprule
Model & Setting & \% Gen. Data & Micro F1 (\%) & PF1 (\%) & EMA (\%) \\
\midrule
Gemma & Cls & 0\%   & 23.97 & --- & --- \\
Gemma & Cls & 75\%  & 32.71 & --- & --- \\
Gemma & Cls & 90\%  & \textbf{35.82} & --- & --- \\
Gemma & Cls & 100\% & 23.95 & --- & --- \\
\midrule
Gemma & ICSF & 0\%   & 20.86 & 10.65 & 0.95 \\
Gemma & ICSF & 75\%  &  30.89 & 33.44 & 8.81 \\
Gemma & ICSF & 90\%  &  \textbf{31.42} & \textbf{35.50} & 8.91 \\
Gemma & ICSF & 100\% &  29.62 & 34.85 & \textbf{9.46} \\
\midrule
LLaMA & Cls & 0\%   & 24.12 & --- & --- \\
LLaMA & Cls & 75\%  &  \textbf{34.53} & --- & --- \\
LLaMA & Cls & 90\%  &  29.50 & --- & --- \\
LLaMA & Cls & 100\% &  27.38 & --- & --- \\
\midrule
LLaMA & ICSF           & 0\%   & 23.70 & 12.25 & 1.12 \\
LLaMA & ICSF           & 75\%  &  30.79 & 32.82 & \textbf{9.43} \\
LLaMA & ICSF           & 90\%  &  \textbf{32.74} & 33.63 & 9.16 \\
LLaMA & ICSF           & 100\% &  31.34 & \textbf{35.38} & 8.71 \\
\bottomrule
\end{tabularx}
}
\caption{Micro F1, PF1 and EMA scores on \acrshort{genbench}'s Test $1_b$ for Gemma-2-9B and LLaMA-3.1-8B across different settings and proportions of generated data.}
\label{tab:gen-test-1b}
\end{table*}

\begin{table*}[t]
\centering
\resizebox{0.65\linewidth}{!}{%
\begin{tabularx}{0.73\linewidth}{l c r c c c}
\toprule
Model & Setting & \% Gen. Data & Micro F1 (\%) & PF1 (\%) & EMA (\%) \\
\midrule
Gemma & Cls & 0\%   & \textbf{55.26} & --- & --- \\
Gemma & Cls & 75\%  & 44.71 & --- & --- \\
Gemma & Cls & 90\%  & 45.59 & --- & --- \\
Gemma & Cls & 100\% & 24.84 & --- & --- \\
\midrule
Gemma & ICSF & 0\%   & 55.66 & 26.73 & 1.41 \\
Gemma & ICSF & 75\%  & 59.16 & 51.41 & 10.42 \\
Gemma & ICSF & 90\%  & \textbf{62.70} & \textbf{54.54} & \textbf{12.16} \\
Gemma & ICSF & 100\% & 53.48 & 49.37 & 9.57 \\
\midrule
LLaMA & Cls & 0\%   & \textbf{55.51} & --- & --- \\
LLaMA & Cls & 75\%  & 47.65 & --- & --- \\
LLaMA & Cls & 90\%  & 38.91 & --- & --- \\
LLaMA & Cls & 100\% & 26.45 & --- & --- \\
\midrule
LLaMA & ICSF           & 0\%   & 55.92 & 27.21 & 1.28 \\
LLaMA & ICSF           & 75\%  & 61.87 & 51.07 & 10.75 \\
LLaMA & ICSF           & 90\%  & \textbf{64.86} & \textbf{51.95} & \textbf{11.33} \\
LLaMA & ICSF           & 100\% & 53.86 & 49.90 & 10.17 \\
\bottomrule
\end{tabularx}
}
\caption{Micro F1, PF1 and EMA scores on \acrshort{genbench}'s Test $2_b$ for Gemma-2-9B and LLaMA-3.1-8B across different settings and proportions of generated data.}
\label{tab:gen-test-2b}
\end{table*}

\begin{table*}[t]
\centering
\resizebox{0.65\linewidth}{!}{%
\begin{tabularx}{0.73\linewidth}{l c r c c c}
\toprule
Model & Setting & \% Gen. Data & Micro F1 (\%) & PF1 (\%) & EMA (\%) \\
\midrule
Gemma & Cls & 0\%   & 73.40 & --- & --- \\
Gemma & Cls & 75\%  & 73.35 & --- & --- \\
Gemma & Cls & 90\%  & 78.53 & --- & --- \\
Gemma & Cls & 100\% & \textbf{78.98} & --- & --- \\
\midrule
Gemma & ICSF & 0\%   & \textbf{66.77} & 8.29 & 0.00 \\
Gemma & ICSF & 75\%  &  64.37 & 50.02 & \textbf{2.85} \\
Gemma & ICSF & 90\%  &  57.94 & \textbf{50.32} & 2.65 \\
Gemma & ICSF & 100\% &  57.73 & 48.02 & 2.35 \\
\midrule
LLaMA & Cls & 0\%   & 72.45 & --- & --- \\
LLaMA & Cls & 75\%  &  70.80 & --- & --- \\
LLaMA & Cls & 90\%  &  72.85 & --- & --- \\
LLaMA & Cls & 100\% &  \textbf{75.09} & --- & --- \\
\midrule
LLaMA & ICSF           & 0\%   & \textbf{64.80} & 8.04 & 0.00 \\
LLaMA & ICSF           & 75\%  &  53.72 & 49.23 & 2.37 \\
LLaMA & ICSF           & 90\%  &  46.96 & 49.41 & 2.32 \\
LLaMA & ICSF           & 100\% &  61.47 & \textbf{50.26} & \textbf{2.67} \\
\bottomrule
\end{tabularx}
}
\caption{Micro F1, PF1 and EMA scores on \acrshort{genbench}'s Test $3_b$ for Gemma-2-9B and LLaMA-3.1-8B across different settings and proportions of generated data.}
\label{tab:gen-test-3b}
\end{table*}

\begin{table*}[t]
\centering
\resizebox{0.65\linewidth}{!}{%
\begin{tabularx}{0.73\linewidth}{l c r c c c}
\toprule
Model & Setting & \% Gen. Data & Micro F1 (\%) & PF1 (\%) & EMA (\%) \\
\midrule
Gemma & Cls & 0\%   &  24.80 & --- & --- \\
Gemma & Cls & 75\%  & 28.51 & --- & --- \\
Gemma & Cls & 90\%  & \textbf{32.57} & --- & --- \\
Gemma & Cls & 100\% & 20.65 & --- & --- \\
\midrule
Gemma & ICSF & 0\%   & 21.29 & 11.39 & 0.87 \\
Gemma & ICSF & 75\%  & 26.90 & 25.79 & \textbf{5.32} \\
Gemma & ICSF & 90\%  & \textbf{27.60} & \textbf{27.69} & 5.00 \\
Gemma & ICSF & 100\% & 25.87 & 26.78 & 5.32 \\
\midrule
LLaMA & Cls & 0\%   & 23.32 & --- & --- \\
LLaMA & Cls & 75\%  & \textbf{30.49} & --- & --- \\
LLaMA & Cls & 90\%  & 24.51 & --- & --- \\
LLaMA & Cls & 100\% & 22.75 & --- & --- \\
\midrule
LLaMA & ICSF           & 0\%   & 22.40 & 12.23 & 1.04 \\
LLaMA & ICSF           & 75\%  & 29.01 & 26.82 & \textbf{5.91} \\
LLaMA & ICSF           & 90\%  & \textbf{30.74} & 27.19 & \textbf{5.91} \\
LLaMA & ICSF           & 100\% & 27.97 & \textbf{28.10} & 5.66 \\
\bottomrule
\end{tabularx}
}
\caption{Micro F1, PF1 and EMA scores on \acrshort{genbench}'s Test $4_b$ for Gemma-2-9B and LLaMA-3.1-8B across different settings and proportions of generated data.}
\label{tab:gen-test-4b}
\end{table*}

Tables~\ref{tab:intent-gen-tests-v2} and \ref{tab:pf1-gen-tests-v2} report the aggregated scores on \acrshort{genbench} for models trained on the $90\%$/$10\%$ and $100\%$/$0\%$ training sets. \Cref{tab:gen-test-1,tab:gen-test-2,tab:gen-test-3,tab:gen-test-4,tab:gen-test-1b,tab:gen-test-2b,tab:gen-test-3b,tab:gen-test-4b} show the the results for Gemma and LLaMa on the individual test cases in \acrshort{genbench}.

\end{document}